\pdfoutput=1
\documentclass{fairmeta}
\usepackage[utf8]{inputenc}
\usepackage{lmodern}
\usepackage{amsmath,amssymb,tabularx}

\title{VisHead-Spec: Head-Routed Visual Attention for Multimodal Speculative Decoding}
\author{Xin Guo}
\author{Arushi Rai}
\author{Xiang Zhou}
\author{Qiang Zhang}
\author{Siyu Lin}
\author{Xiangjun Fan}
\author{Lizhu Zhang}
\affiliation{Meta Recommendation Systems (MRS)}
\correspondence{\email{xinguo@meta.com}}
\hypersetup{pdftitle={VisHead-Spec: Head-Routed Visual Attention for Multimodal Speculative Decoding},pdfauthor={Xin Guo, Arushi Rai, Xiang Zhou, Qiang Zhang, Siyu Lin, Xiangjun Fan, Lizhu Zhang},pdfsubject={Meta Recommendation Systems (MRS)}}
\abstract{
Speculative decoding accelerates vision-language models (VLMs) while preserving the target model's outputs, but its speedup diminishes under high concurrency serving as the system becomes more compute bounded as the batch size increases. This problem is pronounced for multimodal speculative decoding as the long range of the image tokens in the prompt can significantly increase the computational cost during draft token prediction. The block-diffusion drafters (e.g. DFlash, DSpark) can effectively speed up the draft model efficiency by predicting multiple draft tokens in one forward pass, but their speedup for VLMs under high concurrency is still modest (\textless{}9\%). In order to understand this gap, we examine the information flow between different modalities from target model to draft model and identify the sparsity pattern of visual attention between the target and draft models: only small set of key--value heads in the draft model need direct visual feature attention in the context and only visual features from top level layers in the target model need to be attended in the context. We therefore propose VisHead-Spec, which selects a small, fixed set of key--value (KV) heads to attend to all image features from top level layers in the target model and restricts the remaining heads to text features. On Qwen3-VL-8B and MMSpec, VisHead-Spec-8 lets 8 of the drafter's 40 KV heads read the image while keeping accepted length within 0.3\% of the block-diffusion draft model baseline (i.e. DSpark). At batch size 64, its throughput gain over autoregressive decoding increases to 16.0\%, compared with 9.5\% for DSpark; at batch size 1, its speedup remains comparable to DSpark.

}
\begin{document}
\maketitle
\section{Introduction}
\label{sec:source1}

Vision-language models (VLMs) such as Qwen3-VL \citep{bai2025qwen3vl} are now widely used for image captioning, document and chart understanding, and multimodal conversation. Because responses in these tasks often span hundreds of tokens and are generated autoregressively, their generation latency grows in proportion to their length. Speculative decoding \citep{leviathan2023speculativedecoding,chen2023speculativesampling} reduces this latency without altering the output: in each step, a lightweight drafter proposes several future tokens, and the target model evaluates all of them in a single forward pass and accepts the longest prefix that matches its own predictions. However, the speedup of speculative decoding diminishes as more requests are batched together \citep{liu2026performanceillusion}. At small batch sizes, decoding is memory-bound, because each forward pass is dominated by loading the model weights from GPU memory; evaluating several draft tokens in one pass therefore takes about as long as generating a single token \citep{leviathan2023speculativedecoding,liu2026performanceillusion}. At large batch sizes, decoding can become compute-bound, so the additional computation spent on drafting and on verifying tokens that are eventually rejected reduces the speedup provided by speculative decoding. In vLLM \citep{kwon2023pagedattention} on an H100 GPU, for example, the speedup of EAGLE \citep{li2024aeagle} for Llama-3.1-8B on GSM8K falls from 1.73\ensuremath{\times} at batch size 1 to 1.21\ensuremath{\times} at batch size 128 \citep{liu2026performanceillusion}.

In practice, the batch size is set by the latency and throughput requirements of a deployment \citep{kwon2023pagedattention,cheng2026dspark}: services with strict per-user latency targets keep batches small, whereas throughput-oriented services process many requests concurrently to make full use of the GPU. In the production serving of DeepSeek-V4, for example, typical loads involve up to 200 concurrent requests for V4-Flash and 150 for V4-Pro \citep{cheng2026dspark}. For such deployments, whether speculative decoding pays off depends on the cost of each speculative step under concurrent load, which the still-common practice of evaluating at batch size 1 does not capture \citep{liu2026performanceillusion}.

Block-diffusion drafters \citep{chen2026adflash,cheng2026dspark} are among the most effective drafters for text-only language models. Instead of proposing one draft token per forward pass, they fill a block of masked positions in a single pass, so the drafting cost barely grows with the number of draft tokens. They condition on the target model through \emph{KV injection}, in which the hidden states of several layers in the target model are fused into one context feature per position and every draft layer attends to these features as additional keys and values. This design carries over naturally to VLMs, because the hidden states of text tokens in the target model already incorporate visual information \citep{lee2026glance}. However, two challenges arise when block-diffusion drafters are applied to VLMs. 

First, attending to the injected KV context becomes expensive, because it contains every image token and image tokens typically make up most of the prompt. On the text VQA task of the MMSpec benchmark \citep{shen2026mmspec}, for instance, images encoded by Qwen3-VL account for 765 of the 816 prompt tokens on average, and every attention head of every draft layer attends to all of these tokens at every draft step, for every request in the batch. In our A100 TextVQA serving measurements at concurrency 64, drafting accounts for 16.4\% of E2E serving time. Preventing every draft head from attending to image context with the weights held fixed reduces full-batch draft time by 8.10 ms, or 40.8\% of the full-image draft step (Section~\ref{sec:source3-1}).

Second, existing methods that reduce the cost of image tokens in speculative decoding either add computation of their own or lower the accuracy of the drafter. One line of work compresses or selects the visual input of the drafter: ViSpec \citep{kang2025avispec} compresses the image into a few adaptor tokens plus a global feature, SpecVLM \citep{huang2025specvlm} compresses or prunes visual tokens, and TIGER \citep{vo2026tiger} and FOVEA \citep{zhu2026fovea} select a small, state-dependent subset of visual information at each draft step. These methods shorten the visual input that the drafter attends to, but the compression or selection itself requires computation, performed once per image or at every draft step. The other line, represented by HiViS \citep{xie2026hivis}, removes visual tokens from the drafter altogether and relies on the target's fused hidden states. This eliminates the cost of image tokens but lowers the accepted length (Section~\ref{sec:source4-4}). Moreover, all of these methods are designed for autoregressive drafters and are evaluated mainly at batch size 1, whereas the cost of image tokens grows with the number of concurrent requests (Section~\ref{sec:source3-1}).

\begin{figure}[htbp]
\centering
\includegraphics[width=0.90\linewidth,trim=0bp 11bp 0bp 20bp,clip]{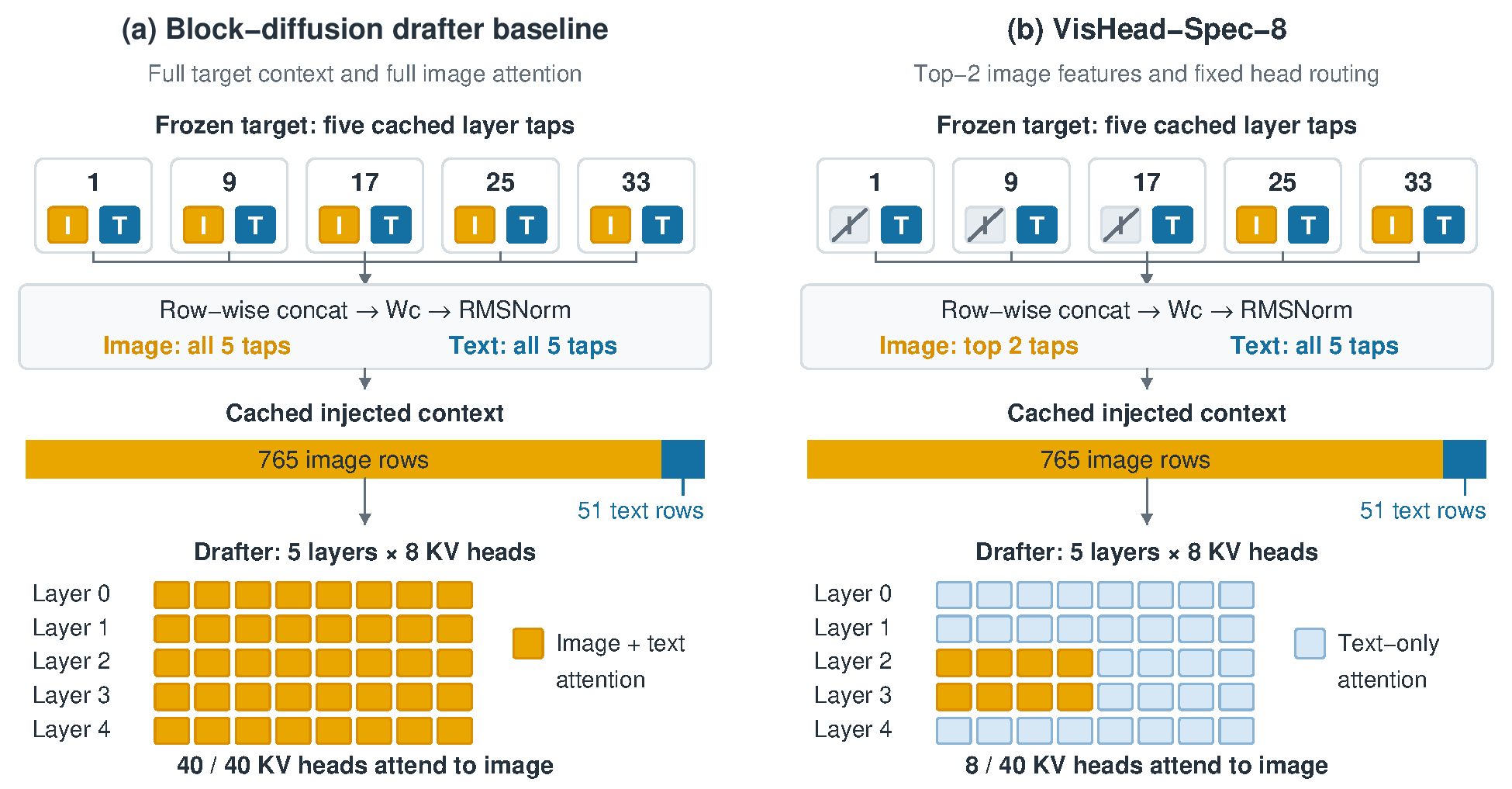}
\caption{Context construction and KV-head routing. Both panels show hidden states cached at layers in the target model S = \{1, 9, 17, 25, 33\} during normal target computation. I and T denote image- and text-token hidden states. (a) DSpark constructs both image and text rows from all five taps and lets all 40 KV heads in its five draft layers attend to image context. (b) VisHead-Spec-8 constructs image rows from the two highest-index sampled layers, 25 and 33, while text rows still use all five. Four KV heads in each of draft layers 2 and 3 retain image access, shown first in each row, and the other heads attend to text context only. Each cell represents one KV head and its four query heads; all heads retain ordinary block attention.}
\label{fig:source1}
\end{figure}

\textbf{VisHead-Spec} (Figure~\ref{fig:source1}) addresses these challenges by restricting image attention at the KV-head level. It constructs image context from 2 sampled layers, 25 and 33 in the target model and retains direct image attention in only 8 of the drafter's 40 KV heads, reducing the number of heads attending to image tokens by 80\%. Every selected head still attends to all image positions, while all heads retain attention to text context and their draft block. The routing mask adds no parameters or per-step token-selection module.

We evaluate VisHead-Spec with Qwen3-VL-8B on the six task categories of MMSpec \citep{shen2026mmspec} at batch sizes from 1 to 64. VisHead-Spec-8 keeps accepted length within 0.3\% of the block-diffusion drafter baseline, DSpark \citep{cheng2026dspark}. At high concurrency (batch size 64), it increases throughput by 16.0\% over autoregressive decoding, compared with 9.5\% for the baseline. At batch size 1, it achieves a 2.23\ensuremath{\times} speedup, comparable to the baseline's 2.26\ensuremath{\times}.

In summary, this paper makes three contributions:

\begin{itemize}
\item[$\bullet$] We identify two forms of sparsity in visual conditioning: image features from the upper sampled layers in the target model retain most of the drafting benefit, and only a small subset of the drafter's KV heads needs direct image attention.
\item[$\bullet$] We introduce VisHead-Spec, which reduces image-context construction and attention costs by selecting image features from 2 sampled layers and a fixed set of KV heads, without adding parameters or changing target verification.
\item[$\bullet$] On Qwen3-VL-8B and MMSpec, VisHead-Spec-8 increases the throughput gain over autoregressive decoding at batch size 64 from 9.5\% to 16.0\%, while remaining comparable to the block-diffusion drafter baseline at batch size 1 (2.23\ensuremath{\times} vs. 2.26\ensuremath{\times}).
\end{itemize}

\section{Related Work}
\label{sec:source2}

\subsection{\textbf{Speculative decoding}}
\label{sec:source2-1}
Speculative decoding verifies a lightweight drafter's proposals with the target while preserving its output distribution \citep{leviathan2023speculativedecoding,chen2023speculativesampling}. Under greedy decoding, only matching proposals are accepted, preserving the target's greedy sequence. EAGLE family \citep{li2024aeagle,li2024beagle2,li2025eagle3} generates proposals autoregressively from target features, but the drafting cost increases with the length of drafted tokens. To address this limitation, block-diffusion based draft models, like DFlash \citep{chen2026adflash}, use a bi-directional attention block to propose multiple draft tokens in one pass. DSpark \citep{cheng2026dspark} adds a lightweight Markov head that conditions block positions on previously sampled tokens and a confidence head that estimates acceptance.

\subsection{\textbf{Speculative decoding for VLMs}}
\label{sec:source2-2}
\citet{gagrani2024multimodalspeculative} showed that a text-only drafter can provide a strong multimodal baseline. Subsequent autoregressive drafters improve visual conditioning: MSD \citep{lin2025msd} separates visual and text tokens and uses two-stage training; DREAM \citep{hu2025dream} combines refined target hidden states through entropy-adaptive cross-attention; HAWK \citep{yang2026hawk} supplies compressed visual hidden states from the target. Other methods shorten visual context. ViSpec \citep{kang2025avispec} compresses image tokens with a vision adapter and adds a global image feature to each text position; SpecVLM \citep{huang2025specvlm} compresses or prunes visual tokens. TIGER \citep{vo2026tiger} and FOVEA \citep{zhu2026fovea} select state-dependent visual tokens or visual-memory entries at each draft step. HiViS \citep{xie2026hivis} removes visual tokens entirely and uses the target's fused text features. These methods use autoregressive drafters and are mostly evaluated at batch size 1. GLANCE \citep{lee2026glance} is closest to our setting: it uses a block-diffusion drafter on Qwen3-VL-8B and finds that reading multimodal context once per block need not add significant overhead at batch size 1, leaving larger-batch serving to future work. 

\section{Method}
\label{sec:source3}

\subsection{Image-attention cost}
\label{sec:source3-1}

\textbf{Image attention is costly under high concurrency.} Every layer of a full-image block-diffusion drafter attends to injected image context at every draft step, so the cost grows with drafter depth and the number of concurrent requests. TextVQA illustrates this effect: images encoded by Qwen3-VL occupy about 94\% of the prompt, or 765 of 816 positions on average. In our A100 serving measurements at batch size 64, drafting accounts for 16.4\% of E2E wall time. With drafter weights held fixed, blocking all attention to image context reduces full-batch draft time by 8.10 ms, or 40.8\% of the full-image draft step. Complete removal also lowers acceptance; Section~\ref{sec:source4-4} compares it with selective image attention.

\subsection{Visual information flow and impacts on speculative decoding}
\label{sec:source3-2}

We study how visual information reaches the drafter and where it is used. We hypothesize that hidden states at text positions in the target model carry useful visual information, but that direct image features remain necessary for parallel draft predictions. We examine three stages: fusion within the target model (Observation 1), construction of the drafter's context (Observations 2--3), and use of that context within the draft block (Observation 4). All layer indices are zero-based.

\begin{figure}[htbp]
\centering
\includegraphics[width=0.76\linewidth,trim=0bp 0.96bp 0bp 4.08bp,clip]{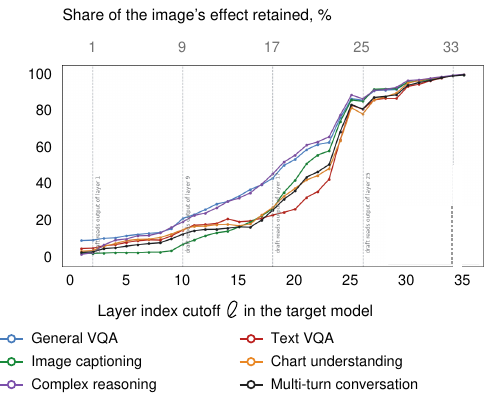}
\caption{Target next-token image-contribution retention on six MMSpec tasks, using Eq.~\eqref{eq:sourceA1}. Text positions cannot attend to image positions at or above layer index cutoff \emph{\ensuremath{\ell}}. Dashed lines mark cutoffs 2, 10, 18, 26, and 34, immediately after the sampled layers 1, 9, 17, 25, and 33 in Qwen3-VL-8B.}
\label{fig:source2}
\end{figure}

\textbf{Observation 1: higher layers in the target model can use visual information already fused into text representations.} We reproduce the qualitative visual-fusion finding of \citet{zhang2025crossmodalflow} on Qwen3-VL-8B-Instruct, using attention knockouts adapted from their protocol on 50 MMSpec test prompts per task. Figure~\ref{fig:source2} shows 99.2--99.6\% retention of the image-related next-token log-probability gain at layer index cutoff \emph{\ensuremath{\ell}} = 34, despite blocking text-to-image attention in layers 34 and 35 of the 36-layer target. This indicates that the text position features at high level layers have already fused sufficient visual information for target model next token prediction. However, the visual information needed for the target model's next-token prediction may differ from that needed by the draft model to predict multiple future tokens. We therefore further examine the visual information flow on the draft model side.

\textbf{Observation 2: text context alone is insufficient for the drafter.} We first generate a fixed greedy reference answer with normal target attention, then feed that answer back to the target to cache hidden states from layers S = \{1, 9, 17, 25, 33\}. A reference token is a token in this fixed answer, not a newly sampled proposal. We score the same reference tokens with the same drafter weights under full image-and-text context and under image-context removal. Text context is identical in both conditions and contains the visual information fused during normal target computation.

In each draft pass, the input contains the last committed token, the anchor \emph{x\textsubscript{t}} at sequence position \emph{t}, followed by five mask embeddings. Every block position can attend to the context at positions before the anchor, subject to the image-attention condition above, and bidirectionally to the embeddings within its own block. To quantify how image-context removal affects these probabilities, we define the reference-token probability ratio \emph{\ensuremath{\rho}}\textsubscript{draft}(\emph{d}) for offsets \emph{d} = 1, \ldots{}, 5:

\begin{equation}
\begin{aligned}
\rho_{\mathrm{draft}}(d)&=\exp\!\Bigg\{
\frac{1}{N}\sum_{r=1}^{N}\frac{1}{|A_r|}\\
&\quad\sum_{t\in A_r}\left[\log q_0(x_{t+d})-\log q_{\mathrm{full}}(x_{t+d})\right]
\Bigg\} .
\end{aligned}
\label{eq:source1}
\end{equation}

Here q\textsubscript{0}(x\textsubscript{t}\textsubscript{+}d) and qfull(x\textsubscript{t}\textsubscript{+}d) are the probabilities of the same reference token under image-context removal and full context at the same anchor. A\textsubscript{r} contains anchors in prompt r with five subsequent answer tokens available, and N is the number of eligible prompts. Eq.~\eqref{eq:source1} averages paired log-probability differences within each prompt and equally across prompts, then exponentiates to obtain a geometric mean probability ratio. A ratio of 1 means no average log-probability change; 0.5 means that reference-token probability is halved on the geometric-mean scale. Multiplication by 100 in the figures expresses the ratio as a percentage.

\begin{figure}[htbp]
\centering
\includegraphics[width=0.90\linewidth,trim=0bp 19.2bp 0bp 3.36bp,clip]{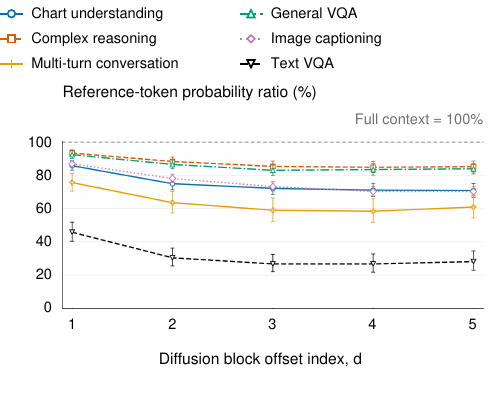}
\caption{Reference-token probability ratios after removing all image-token context from the drafter while holding its weights fixed. Each curve shows one task across diffusion block offsets \emph{d} = 1, \ldots{}, 5, using 100 \ensuremath{\times} \emph{\ensuremath{\rho}\textsubscript{draft}}(\emph{d}) in Eq.~\eqref{eq:source1}. Error bars show paired 95\% confidence intervals across prompts. The dashed horizontal line marks the full-context reference (100\%).}
\label{fig:source3}
\end{figure}

Figure~\ref{fig:source3} shows that text context retains a substantial fraction of the full-context reference-token probability at the first draft offset (d = 1). With all image context removed, the ratio is 75.6--93.4\% on five tasks but only 45.8\% on TextVQA. A possible explanation is that TextVQA requires fine-grained visual details that are harder to recover from text-position representations than the broader scene information needed by tasks such as General VQA.

The probability loss also grows at more distant draft offsets. Between \emph{d} = 1 and \emph{d} = 5, the ratio in Figure~\ref{fig:source3} falls by 8.2--17.6\% across the six tasks. One possible explanation is that the visual evidence fused into the target model's hidden states at text positions is organized for next-token prediction. These hidden states may therefore contain more of the information needed for draft positions close to the anchor than for later positions. 

\begin{figure}[htbp]
\centering
\includegraphics[width=0.76\linewidth,trim=0bp 24bp 0bp 20bp,clip]{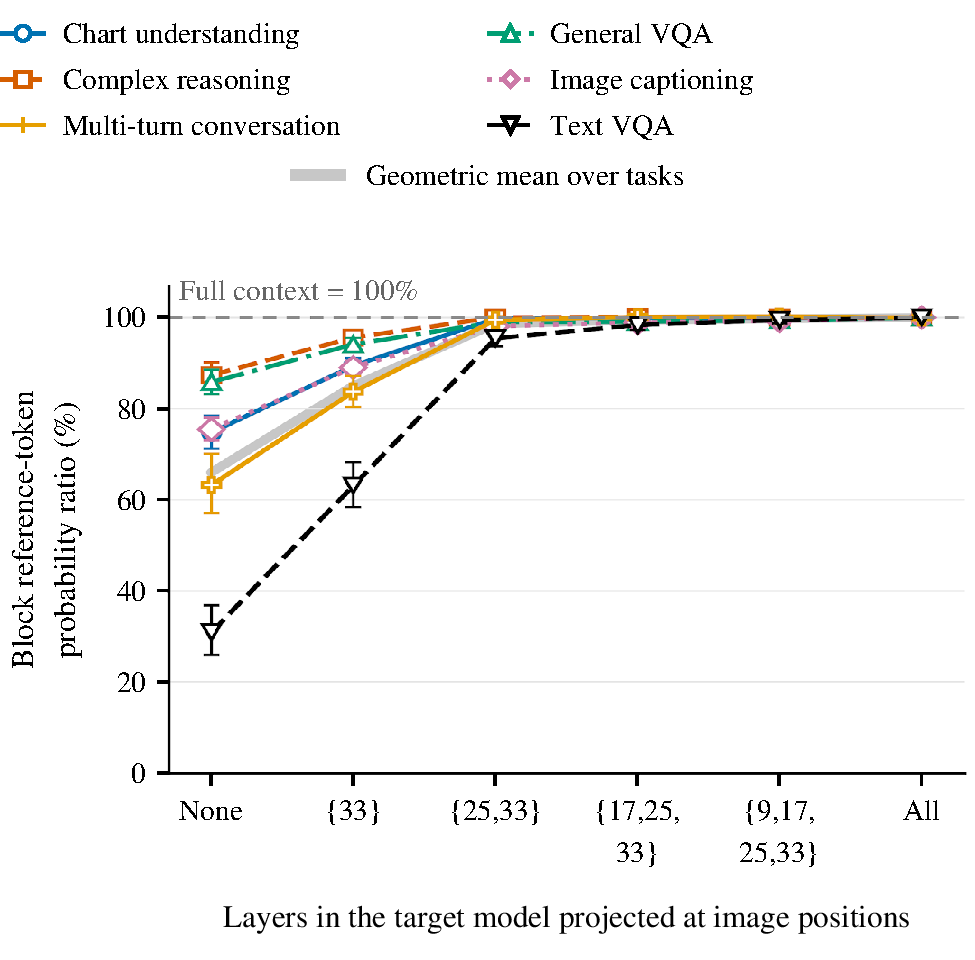}
\caption{Reference-token probability ratios when image rows are constructed from subsets of the sampled layers in the target model. We use Eq.~\eqref{eq:source1} with the subset-conditioned probability in place of \emph{q}\textsubscript{0}, averaged geometrically over the five draft offsets; full context is 100\%. Layers are added in a decreasing index. Tick labels list the selected layers; ``None'' removes image rows, and ``All'' uses \emph{S} = \{1, 9, 17, 25, 33\}. Colored lines show the six tasks and error bars paired 95\% confidence intervals across prompts. Figure~\ref{fig:sourceD1} shows the increasing ordering.}
\label{fig:source4}
\end{figure}

\textbf{Observation 3: the upper layers in the target model provide sufficient image features for drafting.} Observation 2 shows that the draft model needs image context even when its text context contains visual information. We now examine which layers in the target model need to supply those image features. Figure~\ref{fig:source4} shows that image features from layers 25 and 33 recover nearly all of the full-context reference-token probability. The geometric-mean ratio as defined in Eq.~\eqref{eq:source1} over five draft offsets and six tasks is 98.6\% for \{25, 33\}, compared with 66.0\% without any image rows (``None'' in Figure~\ref{fig:source4}). Adding layer 17 raises the ratio only to 99.4\%. TextVQA shows the strongest dependence on image context: its ratio is 30.9\% without image rows and 95.4\% with \{25, 33\}, which indicates that the higher-index sampled layers in the target model provide more of the image information useful to the drafter.

\textbf{Observation 4: only middle layers in the draft model need direct image feature access from the context.} Observation 2 shows that image rows provide information that the frozen drafter does not obtain from text rows, but it does not show in which draft layers the drafter uses this information. Therefore we next examine where the frozen drafter uses image context. We hold its weights and the full injected context from Observation 2 fixed and vary only which draft layers may attend to image context. In each selected layer, all eight KV heads, and hence all 32 query heads, may read image rows. Every layer retains its usual attention to text and block positions. Starting from no image access, we add image-reading layers in increasing index (0, 1, 2, 3, 4) and, separately, in decreasing index (4, 3, 2, 1, 0).

\begin{figure}[htbp]
\centering
\includegraphics[width=0.76\linewidth,trim=0bp 24bp 0bp 20bp,clip]{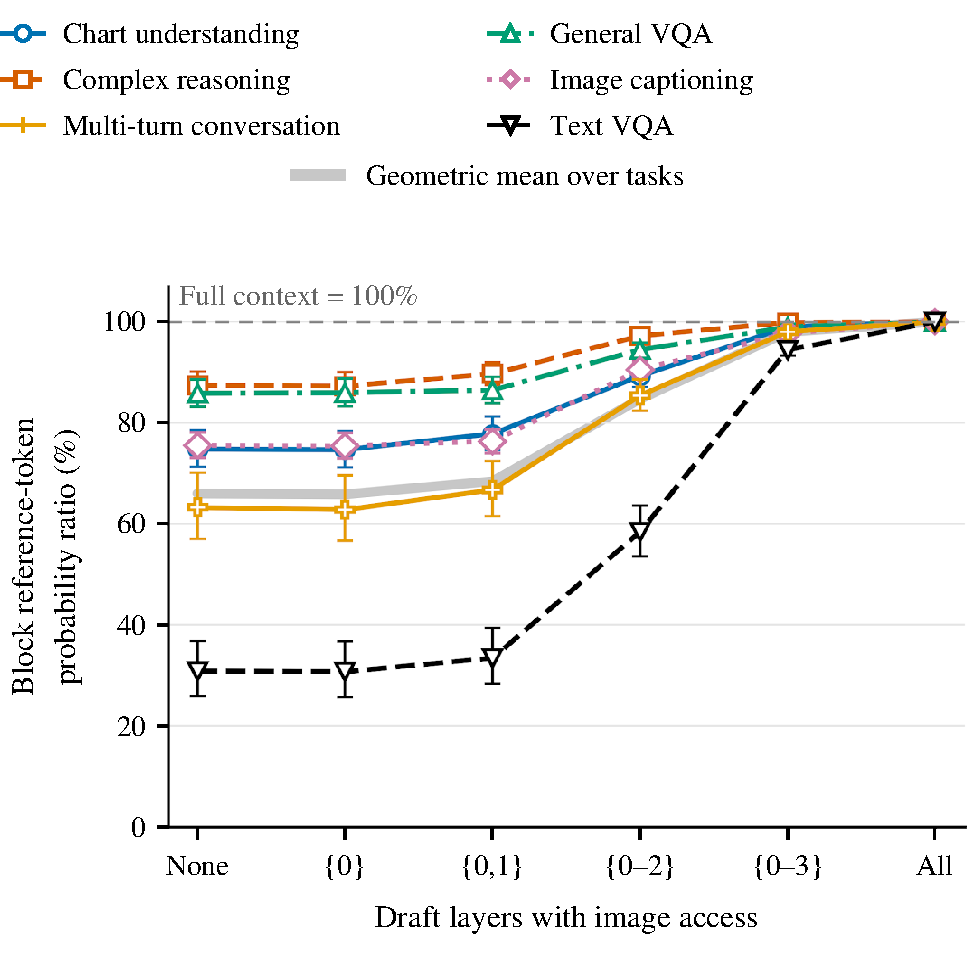}
\caption{Reference-token probability ratios when selected draft layers' KV heads may attend to image rows. Layers gain image access in increasing index; tick labels list the enabled draft layers, ending at full context. The gray line is the geometric mean over six tasks; error bars show paired 95\% confidence intervals across prompts. The reverse ordering is in Figure~\ref{fig:sourceD3}.}
\label{fig:source5}
\end{figure}

The largest improvements occur when draft layers 2 and 3 gain image access. In Figure~\ref{fig:source5}, enabling layers 0 and 1 raises the overall probability ratio from 66.0\% to only 68.4\%, while adding layers 2 and 3 raises it to 98.0\%. The reverse ordering in Figure~\ref{fig:sourceD3} shows the same pattern. Thus, the middle draft layers account for most of the measured benefit of direct image attention. Layer 0 allocates only 8\% of its attention to image positions, while the final layer adds little probability gain once the middle layers can attend to image context.

Together, these observations motivate selecting both the sampled layers that supply image features and the KV heads that attend to them. Section~\ref{sec:source3-3} describes the offline selection procedure; Appendix~\ref{app:sourceB} gives the context construction and allocation details.

\subsection{KV heads routing}
\label{sec:source3-3}

VisHead-Spec constructs image context from layers \{25, 33\} in the target model, while text context uses all five sampled layers \{1, 9, 17, 25, 33\}. A fixed set of KV heads may attend to image context; every head retains ordinary text and within block attention (Figure~\ref{fig:source1}(b)). We select this set offline in two steps.

\textbf{Step 1: selecting draft layers.} We hold the weights of a drafter fine-tuned with full image attention fixed and compare all subsets of its five layers on held-out prompts using offline accepted length (AL), the estimated tokens committed per verification step including the target's bonus token. Layers 2 and 3 retain 81.3\% of the full-image AL gain over no image access, using the normalization in Eq.~\eqref{eq:sourceB2}. The complete comparison is in Appendix~\ref{app:sourceB-2}.

\textbf{Step 2: selecting KV heads.} Using a separate drafter fine-tuned with image access in layers 2 and 3, we hold its weights fixed, rank KV heads within each layer by changes in offline AL when image access is enabled or blocked, and compare head allocations. VisHead-Spec-8 enables four KV heads in each selected layer and VisHead-Spec-4 enables two in each (Appendix~\ref{app:sourceB-3}). We keep the selected mask throughout final fine-tuning and reuse it for every request and decoding step.

We follow the same training setup as DSpark \citep{cheng2026dspark}, with the selected routing mask active throughout training (Appendix~\ref{app:sourceB-4}).

\section{Experiments}
\label{sec:source4}

\subsection{Setup}
\label{sec:source4-1}

\textbf{Target and drafter architecture}. We use the frozen Qwen3-VL-8B-Instruct target \citep{bai2025qwen3vl}, which has 36 layers and hidden size 4096. The VisHead-Spec drafter contains five Qwen3 decoder layers with 32 query heads, 8 KV heads, and head dimension 128, together with a rank-256 Markov head and a confidence head. Its head routing mask restricts direct image access to the selected KV heads. Context injection combines hidden states from layers in the target model \{1, 9, 17, 25, 33\}; the shared embedding table and LM head remain frozen. 

\textbf{Baselines.} We compare against AR, DSpark \citep{cheng2026dspark}, DFlash \citep{chen2026adflash}, EAGLE-3 \citep{li2025eagle3}, MSD \citep{lin2025msd}, and ViSpec \citep{kang2025avispec}. DSpark and DFlash both use a five-layer block decoder. EAGLE-3, MSD, and ViSpec use one-layer autoregressive drafters. For chain verification, EAGLE-3, MSD, and ViSpec make five sequential draft passes and the target evaluates six positions, including the anchor, matching the verification budget of the block drafters. We also evaluate tree verification. EAGLE-3 uses depth 5 with top-8 expansion and 32 verification positions per request. MSD and ViSpec use three draft steps with top-8 expansion and 31 verification positions. These trees trade additional drafting and verification work for a longer accepted prefix.

\textbf{Training initialization.} In the first comparison, VisHead-Spec and DSpark initialize from deepseek-ai/\allowbreak dspark\_\allowbreak qwen3\_\allowbreak 8b\_\allowbreak block7, DFlash from dflash\_qwen3\_8b\_block7, and EAGLE-3 from eagle3\_qwen3\_8b\_ttt7. These released text-only Qwen3-8B checkpoints match the target language model's dimensions and were trained on the same data \citep{yang2025qwen3,cheng2026dspark}. In the separate MSD and ViSpec comparison, all four draft models are randomly initialized; we denote the corresponding block-diffusion draft model baseline and VisHead-Spec models by the suffix ``scratch''.

\textbf{Training data.} We use 67,304 target-answered sequences selected from LLaVA-1.5 mix665k \citep{liu2024allava}, after excluding MMSpec images and questions. These comprise 57,793 image prompts and 9,511 text-only prompts. The comparison groups share this training data corpus. Appendix~\ref{app:sourceB-4} details corpus construction and the training histories.

\textbf{Benchmark tasks.} MMSpec's test split contains 600 prompts, with 100 examples in each of six task categories \citep{shen2026mmspec}. General VQA uses GQA \citep{hudson2019gqa}, and text VQA uses TextVQA \citep{singh2019textvqa}. Image captioning, chart understanding, and complex reasoning use COCO Captions \citep{chen2015cococaptions}, CharXiv \citep{wang2024charxiv}, and MMMU-Pro with ten options \citep{yue2024mmmupro}, respectively. Multi-turn conversation uses ConvBench and MM-MT-Bench \citep{liu2024bconvbench,agrawal2024pixtral}, of which we serve the first turn. Each request contains one image and the MMSpec system prompt.

\textbf{Serving configuration.} Each server runs SGLang 0.5.20 \citep{zheng2024sglang} with our routing implementation on one A100 80GB GPU. We enable CUDA graphs, disable prefix caching, and use greedy decoding with natural end-of-sequence and a maximum of 1,024 new tokens.

\subsection{Overall results}
\label{sec:source4-2}

Table~\ref{tab:source1} compares the AL and throughput speedup against AR for various draft models that are warm initialized from released checkpoints. 

\begin{table}[htbp]
\caption{Mixed-workload AL and E2E throughput speedup relative to AR for drafters initialized from released checkpoints.}
\label{tab:source1}
\centering
\small
\setlength{\tabcolsep}{2pt}
\begin{tabularx}{\columnwidth}{@{}>{\raggedright\arraybackslash}X r r r r r r@{}}
\toprule
\textbf{Method} & \textbf{AL} & \textbf{\shortstack{Batch\\size\\1}} & \textbf{\shortstack{Batch\\size\\8}} & \textbf{\shortstack{Batch\\size\\16}} & \textbf{\shortstack{Batch\\size\\32}} & \textbf{\shortstack{Batch\\size\\64}} \\
\midrule
AR & 1 & 1x & 1x & 1x & 1x & 1x \\
EAGLE-3, chain & 3.00 & 1.85x & 1.59x & 1.36x & 1.00x & 0.89x \\
EAGLE-3, tree & 4.32 & 2.34x & 1.42x & 0.95x & 0.64x & 0.50x \\
DFlash & 3.01 & 2.14x & 1.76x & 1.53x & 1.23x & 1.03x \\
DSpark & 3.28 & 2.26x & 1.86x & 1.62x & 1.29x & 1.09x \\
\textbf{VisHead-Spec-8 (ours)} & \textbf{3.27} & \textbf{2.23x} & \textbf{1.88x} & \textbf{1.65x} & \textbf{1.35x} & \textbf{1.16x} \\
\textbf{VisHead-Spec-4 (ours)} & \textbf{3.26} & \textbf{2.24x} & \textbf{1.87x} & \textbf{1.66x} & \textbf{1.35x} & \textbf{1.16x} \\
\bottomrule
\end{tabularx}
\end{table}

\textbf{Effect of concurrency.} DSpark achieves 2.263\ensuremath{\times} AR throughput at batch size 1, falling to 1.619\ensuremath{\times} at batch size 16 and 1.095\ensuremath{\times} at batch size 64. Other drafters show the same broad decline. As concurrency increases, the additional drafting and verification work consumes more compute, reducing the benefit of committing multiple tokens per speculative step.

\textbf{Throughput improvement under head routing.} VisHead-Spec preserves most of the DSpark baseline's AL across all measured batch sizes. VisHead-Spec-8 lowers AL by only 0.2--0.3\% and VisHead-Spec-4 by 0.5--0.6\%. At batch size 1, VisHead-Spec performs on par with the DSpark baseline. From batch size 8 onward, the measured throughput advantage of VisHead-Spec increases. At batch size 64, VisHead-Spec-8 and VisHead-Spec-4 increase throughput over AR by 16.0\% and 16.5\%, compared with 9.5\% for the DSpark baseline. The throughput gain over AR is therefore about 1.7 times the gain obtained with DSpark.

The two routing variants achieve similar throughput across the measured batches. VisHead-Spec-4 lets four fewer KV heads read image context but reduces AL by another 0.3\% relative to VisHead-Spec-8. The smaller attention cost and lower acceptance largely offset each other, so reducing the routing budget from eight to four gives only a small further throughput change in this setting.

\textbf{Comparison with other initialized baselines.} DFlash has approximately 8\% lower AL than DSpark and is 5--6\% slower across the tested batches, reaching only 1.033\ensuremath{\times} AR throughput at batch size 64. EAGLE-3 chain has similar AL to DFlash and reaches parity with AR at batch size 32 and falls to 0.895\ensuremath{\times} at batch size 64. EAGLE-3 tree obtains the highest AL, 4.32, and the highest batch-1 throughput, exceeding DSpark by 3.6\%. Its 32-position verification becomes costly at larger batches, however, leaving it below AR from batch size 16 onward.

\begin{table}[htbp]
\caption{Mixed-workload AL and E2E throughput speedup relative to AR for randomly initialized drafters.}
\label{tab:source2}
\centering
\small
\setlength{\tabcolsep}{2pt}
\begin{tabularx}{\columnwidth}{@{}>{\raggedright\arraybackslash}X r r r r r r@{}}
\toprule
\textbf{\textbf{Method}} & \textbf{\textbf{AL}} & \textbf{\shortstack{Batch\\size\\1}} & \textbf{\shortstack{Batch\\size\\8}} & \textbf{\shortstack{Batch\\size\\16}} & \textbf{\shortstack{Batch\\size\\32}} & \textbf{\shortstack{Batch\\size\\64}} \\
\midrule
DSpark-scratch & 2.29 & 1.58x & 1.34x & 1.17x & 0.96x & 0.82x \\
MSD, chain & 1.89 & 1.17x & 1.04x & 0.92x & 0.68x & 0.62x \\
MSD, tree & 2.56 & 1.61x & 0.98x & 0.65x & 0.44x & 0.34x \\
ViSpec, chain & 1.82 & 1.09x & 1.00x & 0.88x & 0.64x & 0.60x \\
ViSpec, tree & 2.44 & 1.49x & 0.92x & 0.61x & 0.41x & 0.32x \\
\textbf{VisHead-Spec-8-scratch} & \textbf{2.30} & \textbf{1.58x} & \textbf{1.35x} & \textbf{1.23x} & \textbf{1.01x} & \textbf{0.88x} \\
\bottomrule
\end{tabularx}
\end{table}

\textbf{Results for randomly initialized drafters.} Given there are no released checkpoints for ViSpec, MSD, and DSpark for the same target model and trained with the same data, we train all of these models from scratch with the same training data as mentioned above to make a fair comparison. Table~\ref{tab:source2} shows that VisHead-Spec-8-scratch preserves the AL of DSpark-scratch while increasing throughput by 4.5--7.0\% at batch sizes 16--64. At batch size 32 it reaches 1.012\ensuremath{\times} AR throughput, compared with 0.966\ensuremath{\times} for DSpark-scratch. At batch size 64 both remain below AR, at 0.879\ensuremath{\times} and 0.821\ensuremath{\times}, respectively. MSD and ViSpec chain achieve lower AL, 1.89 and 1.82, and lower throughput than these block drafters. Their trees raise AL to 2.56 and 2.44 and improve batch-1 throughput, but verifying 31 positions leaves both below AR from batch size 8 onward. These results support selective image attention within the controlled DSpark comparison and show the concurrency cost of the tested autoregressive drafters.

\subsection{Results per task}
\label{sec:source4-3}

We next compare different draft models on individual tasks. Table~\ref{tab:source3} combines AL and throughput speedup against AR for drafters initialized from released checkpoints. Throughput is measured at batch size 64, while AL pools per-task runs at batch sizes 8--64. Table~\ref{tab:source2} reports the aggregate AL and throughput comparisons across batch sizes for randomly initialized drafters.

\begin{table}[htbp]
\caption{Per-task results for drafters initialized from released checkpoints. E2E throughput speedup is measured at batch size 64 relative to AR.}
\label{tab:source3}
\centering
\small
\setlength{\tabcolsep}{2pt}
\begin{tabular}{@{}l*{6}{rr}@{}}
\toprule
\raisebox{-1.0ex}{\textbf{Method}} & \multicolumn{2}{c}{\textbf{\shortstack{General\\VQA}}} & \multicolumn{2}{c}{\textbf{\shortstack{Text\\VQA}}} & \multicolumn{2}{c}{\textbf{\shortstack{Image\\captioning}}} & \multicolumn{2}{c}{\textbf{\shortstack{Chart\\understanding}}} & \multicolumn{2}{c}{\textbf{\shortstack{Complex\\reasoning}}} & \multicolumn{2}{c}{\textbf{\shortstack{Multi-turn\\conversation}}} \\
\cmidrule(lr){2-3} \cmidrule(lr){4-5} \cmidrule(lr){6-7} \cmidrule(lr){8-9} \cmidrule(lr){10-11} \cmidrule(lr){12-13}
 & \textbf{AL} & \textbf{Speedup} & \textbf{AL} & \textbf{Speedup} & \textbf{AL} & \textbf{Speedup} & \textbf{AL} & \textbf{Speedup} & \textbf{AL} & \textbf{Speedup} & \textbf{AL} & \textbf{Speedup} \\
\midrule
AR & --- & 1.000 & --- & 1.000 & --- & 1.000 & --- & 1.000 & --- & 1.000 & --- & 1.000 \\
VisHead-8 & 3.056 & \textbf{1.21x} & 3.130 & \textbf{1.15x} & 2.961 & \textbf{1.10x} & 3.628 & \textbf{1.26x} & 3.451 & \textbf{1.18x} & 3.196 & \textbf{1.11x} \\
VisHead-4 & 3.058 & \textbf{1.21x} & 3.110 & \textbf{1.15x} & 2.955 & \textbf{1.11x} & 3.624 & \textbf{1.26x} & 3.445 & \textbf{1.18x} & 3.176 & \textbf{1.12x} \\
DSpark & 3.058 & 1.18x & 3.169 & 1.11x & 2.965 & 1.08x & 3.650 & 1.19x & 3.452 & 1.13x & 3.199 & 1.00x \\
DFlash & 2.807 & 1.13x & 2.903 & 1.06x & 2.709 & 1.01x & 3.377 & 1.14x & 3.182 & 1.06x & 2.933 & 0.96x \\
EAGLE-3 chain & 2.836 & 1.01x & 2.832 & 0.98x & 2.714 & 0.89x & 3.345 & 1.06x & 3.192 & 0.93x & 2.917 & 0.86x \\
EAGLE-3 tree & 4.193 & 0.54x & 4.093 & 0.60x & 4.130 & 0.44x & 4.615 & 0.54x & 4.473 & 0.47x & 4.198 & 0.53x \\
\bottomrule
\end{tabular}
\end{table}

\textbf{Largest and smallest task gains.} At batch size 64, both VisHead-Spec variants outperform the block-diffusion drafter baseline on all six tasks. Multi-turn conversation has the largest improvements: 10.38\% for VisHead-Spec-8 and 11.34\% for VisHead-Spec-4. The smallest improvements are 1.02\% for VisHead-Spec-8 on captioning and 1.79\% for VisHead-Spec-4 on General VQA. With a fixed head budget, a larger fraction of image positions in the injected context permits a larger reduction in context-attention work. E2E gains also depend on accepted length, answer length, and target verification cost, so image-token fraction alone does not determine the throughput improvement.

\subsection{Ablation study}
\label{sec:source4-4}

Image-feature selection during draft prefill. We compare constructing image rows from all five sampled layers in the target model \{1, 9, 17, 25, 33\} with the default two-layer subset \{25, 33\}, while retaining all five layers for text rows. In the recorded A100 draft-prefill microbenchmark, image positions account for 93\% of prompt tokens. Selecting two layers reduces the input width of Wc at image positions from 5 \ensuremath{\times} 4096 to 2 \ensuremath{\times} 4096. Aggregated over the measured prefill batches, context-projection time decreases from 286.4 to 143.5 ms, and total context-construction time decreases from 498.3 to 360.5 ms, a 27.7\% reduction.

\textbf{Selective image access versus complete removal and dynamic routing.} We compare VisHead-Spec-0 (complete image row removal from context), VisHead-Spec-8, and VisHead-Spec-4 after fine-tuning each from the same released checkpoint on the same data and schedule. Table~\ref{tab:source4} reports mixed-workload AL and throughput speedup. 

\begin{table}[htbp]
\caption{Mixed-workload ablation of image-attention budgets after matched fine-tuning. AL is averaged over batch sizes 1, 8, 16, 32, and 64; each batch column is E2E throughput divided by AR throughput at that batch size.}
\label{tab:source4}
\centering
\small
\setlength{\tabcolsep}{2pt}
\begin{tabularx}{\columnwidth}{@{}>{\raggedright\arraybackslash}X r r r r r r@{}}
\toprule
\textbf{\textbf{Variant}} & \textbf{\textbf{AL}} & \textbf{\shortstack{Batch\\size\\1\\speedup}} & \textbf{\shortstack{Batch\\size\\8\\speedup}} & \textbf{\shortstack{Batch\\size\\16\\speedup}} & \textbf{\shortstack{Batch\\size\\32\\speedup}} & \textbf{\shortstack{Batch\\size\\64\\speedup}} \\
\midrule
VisHead-Spec-0 & 3.152 & 2.19x & 1.83x & 1.62x & 1.31x & 1.12x \\
VisHead-Spec-8 (ours) & 3.27 & 2.23x & 1.88x & 1.65x & 1.35x & 1.16x \\
VisHead-Spec-4 (ours) & 3.26 & 2.24x & 1.87x & 1.66x & 1.35x & 1.16x \\
\bottomrule
\end{tabularx}
\end{table}

Complete image removal reduces mean AL to 3.152, compared with 3.27 and 3.26 for the eight- and four-head variants. VisHead-Spec-0 consequently remains slower than both routed variants at every measured batch, despite reading fewer context rows. At batch size 64 it reaches 1.127\ensuremath{\times} AR throughput, compared with 1.160\ensuremath{\times} and 1.165\ensuremath{\times} for VisHead-Spec-8 and VisHead-Spec-4. Selective image access therefore provides a better acceptance--cost trade-off than removing all image context in these runs. 

We also compare the static mask with a trained per-token router. The router lowers accepted length and throughput relative to always-on image access through the same four KV heads in the measured batch-1 setting (Appendix~\ref{app:sourceE}).

\textbf{VisHead-Spec on DFlash.} Figure~\ref{fig:sourceD4} reports the DFlash routing ablation at a server batch cap of 64, using the attention-serving implementation of the block-diffusion drafter baseline and all five sampled context layers. Mean AL is 2.925 with full image attention, compared with 2.906 and 2.891 for four- and eight-head routing. Relative to full-image DFlash, paired E2E throughput changes are +6.78 \ensuremath{\pm} 1.89\% and +5.42 \ensuremath{\pm} 0.69\%, respectively. These are 95\% Student-t intervals over three paired execution orders. Appendix~\ref{app:sourceA-3} specifies the measurement protocol.

\section{Conclusion}
\label{sec:source5}

Block-diffusion drafters avoid sequential draft passes, but their repeated attention to image-heavy context can limit speculative speedup under concurrent VLM serving. Our study shows that text representations contain useful visual information, but removing image-token context from the draft model can hurt the speedup performance. A further investigation shows that a small set of KV heads preserve much of the acceptance gained from direct image attention. Based on these observations, we proposed VisHead-Spec, which can retain image attention for those heads and restricts the others to text context. On Qwen3-VL-8B and MMSpec, VisHead-Spec-8 keeps AL within 0.3\% of DSpark and increases the throughput gain over AR at batch size 64 from 9.5\% to 16.0\%, while remaining comparable to DSpark at batch size 1. 

\clearpage
\nocite{*}
\bibliographystyle{icml2026}
\bibliography{references}
\section*{Impact Statement}
\label{sec:impact-statement}

This work aims to improve the inference efficiency of vision-language models. More efficient inference may reduce compute costs and enable wider deployment, while retaining the target model's outputs. The potential societal consequences of wider deployment, including inaccurate or biased outputs, remain relevant; deployment should preserve the safeguards applied to the target model.

\clearpage
\appendix
\makeatletter
\@addtoreset{figure}{section}
\@addtoreset{table}{section}
\@addtoreset{equation}{section}
\makeatother
\renewcommand{\thefigure}{\thesection\arabic{figure}}
\renewcommand{\thetable}{\thesection\arabic{table}}
\renewcommand{\theHfigure}{appendix.\thesection.\arabic{figure}}
\renewcommand{\theHtable}{appendix.\thesection.\arabic{table}}
\setcounter{equation}{0}
\renewcommand{\theequation}{\thesection\arabic{equation}}
\renewcommand{\theHequation}{appendix.\thesection.\arabic{equation}}

\section{Evaluation and Profiling Protocols}
\label{app:sourceA}

\subsection{Attention knockouts in the target model}
\label{app:sourceA-1}

We evaluate Qwen3-VL-8B-Instruct on 50 MMSpec test prompts per task using attention knockouts adapted from \citet{zhang2025crossmodalflow}. Their LLaVA analysis blocks attention in sliding windows of nine layers; ours uses a cumulative layer index cutoff and evaluates fixed reference-answer tokens. The target first generates one greedy answer with normal image attention. We feed this fixed answer back and score the same reference tokens under three conditions: normal attention in all layers; a complete text-to-image knockout; and a partial knockout at and above a layer index cutoff \emph{\ensuremath{\ell}}.

\textbf{Blocking attention to image tokens.} In Observation 1 (Section~\ref{sec:source3-2}), selected text-to-image attention scores are set to \ensuremath{-}\ensuremath{\infty}, following \citet[Eq.~7]{zhang2025crossmodalflow}. The layer index cutoff \emph{\ensuremath{\ell}} is the first layer in the target model where all text positions are prevented from reading image keys and values; layers with indices below the cutoff retain normal attention, while layers with indices at or above it use the restriction. The complete knockout applies this restriction in every layer while processing the prompt and the same reference answer. The target's LM head scores the token immediately following each predecessor, as specified in Eq.~\eqref{eq:sourceA1}.

We quantify how much of the target's image-related log-probability gain remains after the partial knockout using the retention score \emph{Ret\textsubscript{next}}(\emph{\ensuremath{\ell}}):

\begin{equation}
\begin{aligned}
\operatorname{Ret}_{\mathrm{next}}(\ell)
&=\frac{1}{N}\sum_{r=1}^{N}\\
&\quad
\frac{\sum_{j\in D_r}(\log p_{\ell,j}-\log p_{\mathrm{hidden},j})}
{\sum_{j\in D_r}(\log p_{\mathrm{full},j}-\log p_{\mathrm{hidden},j})} .
\end{aligned}
\label{eq:sourceA1}
\end{equation}

For reference token \emph{x\textsubscript{j}}, let \emph{p\textsubscript{full,j}}, \emph{p\textsubscript{hidden,j}}, and \emph{p\textsubscript{\ensuremath{\ell},j}} be its probabilities under normal attention, the complete knockout, and the partial knockout at layer index cutoff \emph{\ensuremath{\ell}}, respectively. Within each prompt \emph{r}, Eq.~\eqref{eq:sourceA1} sums the retained and full-image log-probability gains over image-dependent answer positions \emph{D\textsubscript{r}}, then takes their ratio. It averages these ratios equally across the \emph{N} eligible prompts. Appendix~\ref{app:sourceA} defines the image-dependence filter that is used to define \emph{D\textsubscript{r}}. Expressed as a percentage, 100\% of retention score \emph{Ret\textsubscript{next}}(\emph{\ensuremath{\ell}}) means that the partial knockout retains the full-image gain, while 0\% means that no gain remains over the complete knockout.

\textbf{Image-dependent answer tokens.} We define \emph{D\textsubscript{r}} using positions where the normal target pass predicts the reference answer token as its argmax. A position is retained if the complete text-to-image knockout changes that argmax or decreases the same token's log probability by at least 1, corresponding to a probability drop of about 63\%. This filter retains 8--27\% of answer positions depending on the task and is determined before applying Eq.~\eqref{eq:sourceA1}. All log probabilities refer to that fixed reference answer token under the specified attention condition.

\subsection{Context comparisons in the drafter}
\label{app:sourceA-2}

The drafter comparisons use Eq.~\eqref{eq:source1} on all valid reference tokens at complete five-token anchors, without the target's image-dependence filter. Their paired log-probability differences are averaged within prompts and then equally across prompts before exponentiation. These ratios compare complete image-context removal with the same frozen drafter's full context; both conditions retain the same text context vectors formed from the normally image-conditioned hidden states in the target model and the same reference predecessors. Target next-token retention and drafter probability ratios are different diagnostics; neither alone determines multi-token accepted length, which depends on sequential verification.

\subsection{DFlash routing measurement protocol}
\label{app:sourceA-3}

The DFlash ablation uses 600 fixed MMSpec requests (100 per task), server batch cap 64, a maximum of 1,024 new tokens, and three paired execution orders. All conditions retain the same attention implementation and all five sampled context layers. Throughput is completion tokens divided by complete benchmark wall time; paired changes are 100 \ensuremath{\times} (TPSrouted/TPSfull \ensuremath{-} 1), with 95\% Student-t intervals (two degrees of freedom). AL is the request-mean completion-token/verification-pass ratio, including the initial prefill token. Fixed target-answer replay measures controlled serving cost; it does not measure on-policy generation quality or training-seed variance.

\FloatBarrier
\vspace{1em}
\section{Selection and Training Details}
\label{app:sourceB}

\subsection{Context construction and head routing}
\label{app:sourceB-1}

\textbf{Selecting layers in the target model.} For context position \emph{j}, let \emph{h\textsubscript{j}}\textsuperscript{(\ensuremath{\ell})} denote the frozen target's hidden state at layer \ensuremath{\ell}. We concatenate states from sampled layers \emph{S} = \{1, 9, 17, 25, 33\}, then apply a learned projection \emph{W\textsubscript{c}} and RMS normalization to obtain a context row:

\begin{equation}
\begin{aligned}
c_j &= \operatorname{RMSNorm}\!\left(W_c\,\operatorname{concat}_{\ell\in S}h_j^{(\ell)}\right)
\in\mathbb{R}^{4096} .
\end{aligned}
\label{eq:sourceB1}
\end{equation}

We cache \emph{c\textsubscript{j}} for subsequent draft steps. Let \emph{I} denote image positions. By default, VisHead-Spec constructs image rows from layers \{25, 33\} in the target model, the two highest-index sampled layers studied in Observation 3. For \emph{j} \ensuremath{\in} \emph{I}, Eq.~\eqref{eq:sourceB1} uses only the corresponding columns of \emph{W\textsubscript{c}}. Text rows continue to use all five sampled layers.

\textbf{Head routing.} VisHead-Spec gives a small, fixed set of KV heads access to image context, while the remaining heads attend only to text context (Figure~\ref{fig:source1}(b)). We choose this set offline through two comparisons: first selecting draft layers that preserve acceptance with image access, then allocating KV heads within those layers. Each comparison holds an initially trained drafter's weights fixed; the corresponding checkpoint and data are specified below. We retain the selected mask throughout the final variants' fine-tuning and reuse it for every request and decoding step.

The routing set \emph{\ensuremath{\mathcal{R}}} contains the selected (draft layer, KV head) pairs. Each draft layer \emph{i} uses grouped-query attention \citep{ainslie2023gqa} with 32 query heads sharing eight KV heads. Query head \emph{q} \ensuremath{\in} \{0, \ldots{}, 31\} uses KV head \emph{g}(\emph{q}) = \ensuremath{\lfloor}\emph{q}/4\ensuremath{\rfloor}. There are two context cases: every head can attend to eligible text positions before the anchor, while image positions are visible only when (\emph{i}, \emph{g}(\emph{q})) belongs to \emph{\ensuremath{\mathcal{R}}}. All query heads sharing one KV head also share the same image-access decision, making the KV head the unit of the image-attention budget. The full-image baseline enables all 40 KV heads across five draft layers, while VisHead-Spec-8 and VisHead-Spec-4 enable 8 and 4, respectively. These counts refer to heads across all draft layers, rather than to a subset of image tokens.

\subsection{Draft-layer selection}
\label{app:sourceB-2}

\textbf{Routing strategy selection.} We select layers first, then allocate KV heads within the selected layers. At each reference answer anchor, we score five proposals against the same target-generated answer, count the consecutive argmax matches before the first mismatch, and add one for the target's bonus token. We average this count over anchors within each prompt and then equally over prompts to obtain offline AL, \emph{\ensuremath{\tau}}, the estimated mean number of tokens committed per verification step.

\textbf{Stage 1: selecting layers.} We use a drafter already fine-tuned for one epoch with full image attention, hold its weights fixed, and evaluate 100 held-out prompts. For each prompt, we cache a normally generated target answer and its context features, then score that answer with all five draft layers reading image context, with no layer reading it, and with only a candidate subset \emph{U} reading it. Let \emph{\ensuremath{\tau}\textsubscript{all}}, \emph{\ensuremath{\tau}\textsubscript{none}}, and \emph{\ensuremath{\tau}\textsubscript{U}} be the corresponding offline AL values. We measure the fraction of the full-image AL gain retained by \emph{U}:

\begin{equation}
\operatorname{retained}(U)=
\frac{\tau_U-\tau_{\mathrm{none}}}{\tau_{\mathrm{all}}-\tau_{\mathrm{none}}} .
\label{eq:sourceB2}
\end{equation}

We evaluate all 32 subsets of draft layers \{0, 1, 2, 3, 4\}. Table~\ref{tab:sourceB1} reports all five single-layer candidates and the three candidates with the highest offline AL at each budget of two, three, and four layers. Appendix~\ref{app:sourceB-4} gives the training configuration.

\begin{table}[htbp]
\caption{Layer subsets from the complete 32-subset comparison, including every single-layer candidate and the best three at each larger budget. AL gain retained is Eq.~\eqref{eq:sourceB2} expressed as a percentage, measured on 100 held-out prompts with the same drafter weights and target context in every condition.}
\label{tab:sourceB1}
\centering
\small
\begin{tabular}{@{}lr@{}}
\toprule
\textbf{\textbf{Layers with image attention}} & \textbf{\textbf{AL gain retained (\%)}} \\
\midrule
0 & 0.1 \\
1 & 3.4 \\
2 & 29.6 \\
3 & 41.2 \\
4 & 10.6 \\
2, 3 & 81.3 \\
1, 3 & 57.2 \\
2, 4 & 53.4 \\
2, 3, 4 & 90.1 \\
1, 2, 3 & 90.0 \\
0, 2, 3 & 82.1 \\
1, 2, 3, 4 & 98.9 \\
0, 1, 2, 3 & 91.9 \\
0, 2, 3, 4 & 90.8 \\
\bottomrule
\end{tabular}
\end{table}

The layer comparison identifies layers 2 and 3 for subsequent head selection. Layer 3, the best single layer, retains 41.2\% of the full-image AL gain, while the pair \{2, 3\} retains 81.3\%. The best three- and four-layer subsets retain 90.1\% and 98.9\%, respectively, but require more heads to read image context. We also test other layer selection methods like image token attention weight based selection. However selection based on image attention weight does not reliably identify the layers that preserve acceptance: layer 4 accounts for the largest share of attention on image positions, 27\%, yet retains only 10.6\% of the AL gain when enabled alone.

\subsection{KV-head selection}
\label{app:sourceB-3}

\textbf{Stage 2: selecting KV heads.} We use a separate drafter fine-tuned for two epochs with all eight KV heads in each of layers 2 and 3 reading image context; its other layers attend only to text context. We hold its weights fixed and evaluate 200 holdout prompts. Every candidate uses the same cached target answers, context features, and offline AL metric. The controls enable all 16 heads in the two selected layers or disable image attention entirely.

We use different reference masks to rank heads in the two layers. For layer 2, we begin with all layer-3 heads enabled and all layer-2 heads blocked, then enable one layer-2 head at a time and measure its AL gain. This asks which layer-2 heads add useful image access beyond the strongest single layer. For layer 3, we begin with all 16 heads enabled, then block one layer-3 head at a time and measure the AL loss, testing which layer-3 heads the complete two-layer mask relies on. Thus the layer-2 scores measure additions to the layer-3 reference, while the layer-3 scores measure removals from the full two-layer reference. They are conditional contributions used to rank heads within each layer, not directly comparable scores across layers.  Writing H = \{0, \ldots{}, 7\} and \ensuremath{\tau}(U\textsubscript{2}, U\textsubscript{3}) for offline AL with enabled head sets U\textsubscript{2} and U\textsubscript{3} in draft layers 2 and 3, the scores are s\textsubscript{2}(h) = \ensuremath{\tau}(\{h\}, H) \ensuremath{-} \ensuremath{\tau}(\ensuremath{\varnothing}, H) and s\textsubscript{3}(h) = \ensuremath{\tau}(H, H) \ensuremath{-} \ensuremath{\tau}(H, H\ensuremath{\setminus}\{h\}). All other draft layers have image attention blocked in these comparisons.

\begin{table}[htbp]
\caption{Offline AL for selected allocations from the 81-allocation comparison. ``Best at budget'' identifies the highest measured AL at the given total head count.}
\label{tab:sourceB2}
\centering
\small
\setlength{\tabcolsep}{4pt}
\begin{tabularx}{\textwidth}{@{}>{\raggedright\arraybackslash}p{0.18\textwidth} >{\centering\arraybackslash}p{0.18\textwidth} c >{\raggedright\arraybackslash}X@{}}
\toprule
\textbf{\textbf{KV heads attending to image context}} & \textbf{\textbf{Head count for Layer 2 + layer 3}} & \textbf{\textbf{Offline AL}} & \textbf{\textbf{Note}} \\
\midrule
16 & 8 + 8 & 3.32 & All heads in layers 2 and 3 \\
12 & 8 + 4 & 3.32 & Best at budget 12 \\
10 & 6 + 4 & 3.31 & Best at budget 10 \\
8 & 4 + 4 & 3.29 & VisHead-Spec-8 \\
8 & 6 + 2 & 3.29 & Best at budget 8 \\
6 & 3 + 3 & 3.27 & Best at budget 6 \\
4 & 2 + 2 & 3.25 & VisHead-Spec-4; best at budget 4 \\
4 & 0 + 4 & 3.10 & Four heads in layer 3 \\
4 & 4 + 0 & 3.06 & Four heads in layer 2 \\
2 & 1 + 1 & 3.16 & Best at budget 2 \\
0 & 0 + 0 & 2.87 & VisHead-Spec-0 (No image attention) \\
\bottomrule
\end{tabularx}
\end{table}

Table~\ref{tab:sourceB2} shows that a small head budget preserves most of the acceptance benefit of image attention. Offline AL changes from 3.328 with all 16 heads budget to 3.293 with the balanced eight-head budget and 3.250 with the four-head budget, compared with 2.873 with no image budget. At the same four-head budget, splitting the heads 2 + 2 across the layers gives higher AL than placing all four in layer 3 (3.099) or layer 2 (3.057). Both the budget and its distribution across layers therefore affect acceptance.

\subsection{Training}
\label{app:sourceB-4}

We train the block decoder, Markov head, and confidence head jointly while keeping the target, shared embedding table, and LM head frozen. The selected routing mask is active throughout training, giving each head the same context access that it will have at serving time. Following \citet{cheng2026dspark}, we combine token prediction, alignment to the target distribution, and confidence prediction in the loss:

\begin{equation}
\mathcal{L}=0.1\,\mathcal{L}_{\mathrm{CE}}
+0.9\,\mathcal{L}_{\mathrm{L1}}+1.0\,\mathcal{L}_{\mathrm{conf}} .
\label{eq:sourceB3}
\end{equation}

At a draft position with reference token y, let p\textsuperscript{d} and p\textsuperscript{t} be the draft and target distributions over the shared vocabulary and c be the predicted confidence. The per-position losses are \ensuremath{\ell}CE = \ensuremath{-}log p\textsuperscript{d}(y), \ensuremath{\ell}L1 = \ensuremath{\Sigma}v |p\textsuperscript{d}(v) \ensuremath{-} p\textsuperscript{t}(v)|, and \ensuremath{\ell}conf = \ensuremath{-}a log c \ensuremath{-} (1\ensuremath{-}a) log(1\ensuremath{-}c), where a = 1 \ensuremath{-} 0.5 \ensuremath{\Sigma}v |p\textsuperscript{d}(v) \ensuremath{-} p\textsuperscript{t}(v)| is a detached soft acceptance target. Following \citet{cheng2026dspark}, each \ensuremath{\mathcal{L}} term in Eq.~\eqref{eq:sourceB3} is the normalized weighted average of its per-position loss, with weights exp(\ensuremath{-}d/4) over draft offsets d. The confidence loss therefore uses binary cross-entropy with a soft target, rather than a label indicating whether the argmax tokens match.

Routing is implemented through a per-head block mask. In each draft layer, block positions attend to the fixed context before the anchor and bidirectionally within the block. KV heads selected by \ensuremath{\mathcal{R}} may also attend to image context rows, while the remaining heads use the same pattern with image context rows hidden. Section~\ref{sec:source3-3} specifies which heads receive image access. Since only the mask changes, this implementation adds no parameters.

\begin{table}[htbp]
\caption{Data, initialization, objectives, and schedules for drafters initialized from released checkpoints. VisHead-Spec variants differ in their routing masks. The separately initialized MSD and ViSpec comparison follows Table~\ref{tab:sourceB4}.}
\label{tab:sourceB3}
\centering
\small
\setlength{\tabcolsep}{4pt}
\begin{tabularx}{\textwidth}{@{}>{\raggedright\arraybackslash}p{0.26\textwidth} >{\raggedright\arraybackslash}X@{}}
\toprule
\textbf{Item} & \textbf{Setting} \\
\midrule
Target & Qwen3-VL-8B-Instruct (frozen; 36 layers, hidden size 4096) \\
Drafter (VisHead-Spec and block-diffusion drafter baseline) & 5 Qwen3 decoder layers, 32 query / 8 KV heads, head dimension 128; block size 7; Markov head of rank 256; confidence head; context projection of hidden states from layers 1, 9, 17, 25, 33 in the target model; embedding and LM head shared with the target and frozen \\
Training warm start & deepseek-ai/dspark\_qwen3\_8b\_block7 (VisHead-Spec and block-diffusion drafter baseline), dflash\_qwen3\_8b\_block7 (DFlash), eagle3\_qwen3\_8b\_ttt7 (EAGLE-3) \\
Training data & 67,304 sequences: 68,000 first-turn prompts from LLaVA-1.5 mix665k (first answer \ensuremath{\geq} 5 words; MMSpec images and questions removed; about 15\% text-only ShareGPT), re-answered greedily by the target at native resolution (\ensuremath{\leq} 2,048 tokens; looping answers dropped) \\
Schedule & 3 epochs = 8,412 steps; global batch 24; AdamW, learning rate 4\ensuremath{\times}10\textsuperscript{-}\textsuperscript{5}, 4\% warmup, cosine decay to 0; gradient clipping 1.0; bf16; AMD MI300X; identical sample order for all drafters; final checkpoint \\
Head router & per-head block mask; image token id of Qwen3-VL; no added parameters \\
\bottomrule
\end{tabularx}
\end{table}

\begin{table}[htbp]
\caption{Architectures and method-specific training schedules for the randomly initialized comparison. All methods use the same 67,304-sequence corpus and a nominal ten-epoch data budget. The target is frozen throughout.}
\label{tab:sourceB4}
\centering
\small
\setlength{\tabcolsep}{4pt}
\begin{tabularx}{\textwidth}{@{}>{\raggedright\arraybackslash}p{0.16\textwidth} >{\raggedright\arraybackslash}X >{\raggedright\arraybackslash}X >{\raggedright\arraybackslash}p{0.22\textwidth}@{}}
\toprule
\textbf{\textbf{Method}} & \textbf{\textbf{Architecture}} & \textbf{\textbf{Training schedule}} & \textbf{\textbf{Optimizer steps / data}} \\
\midrule
DSpark-scratch & 5 layers; 32 Q / 8 KV heads per layer; all 40 KV heads read images & AdamW; LR 6\ensuremath{\times}10\textsuperscript{-}\textsuperscript{4}; global batch 512; 4\% warmup, cosine decay & 10 epochs; 1,310 optimizer steps \\
VisHead-Spec-8-scratch & Same decoder as DSpark; 8 image-reading KV heads in layers 2 and 3 & Same schedule as DSpark-scratch & 10 epochs; 1,310 optimizer steps \\
MSD & 1 layer; 32 Q / 8 KV heads; final target hidden state plus token/visual embedding & AdamW; LR 5\ensuremath{\times}10\textsuperscript{-}\textsuperscript{5}; global batch 32; 2,000 warmup steps; norm clipping; paper objective & 21,030 steps; text-first curriculum: 63k text draws, 63k draws ramping image fraction 0\ensuremath{\to}1, then image data \\
ViSpec & 1 layer; 32 KV heads; image adaptor with 2 queries, one retained image row and one global feature & Stage 1: AdamW 3\ensuremath{\times}10\textsuperscript{-}\textsuperscript{5} on text. Stage 2: AdamW 3\ensuremath{\times}10\textsuperscript{-}\textsuperscript{6} on image data; global batch 8; MTP 1 & Stage 1: 9,511 text rows \ensuremath{\times}10. Stage 2: 57,793 image rows \ensuremath{\times}10; 72,240 image-stage steps \\
\bottomrule
\end{tabularx}
\end{table}

\FloatBarrier
\vspace{1em}
\section{Per-Task Results at All Batch Sizes}
\label{app:sourceC}

Tables~\ref{tab:sourceC1}--\ref{tab:sourceC5} extend the per-task comparison to every measured batch size. Each entry is the E2E throughput change relative to DSpark, expressed as a percentage with mean \ensuremath{\pm} standard error. Batch sizes 8--64 use separate runs for each task, while batch size 1 uses request service times from the mixed runs. Method abbreviations follow Table~\ref{tab:source3}.

\begin{table}[htbp]
\caption{Per-task E2E throughput change relative to DSpark at batch size 64.}
\label{tab:sourceC1}
\centering
\small
\setlength{\tabcolsep}{4pt}
\begin{tabularx}{\textwidth}{@{}>{\raggedright\arraybackslash}X r r r r r r@{}}
\toprule
\textbf{Task} & \textbf{VisHead-0} & \textbf{VisHead-8} & \textbf{VisHead-4} & \textbf{DFlash} & \textbf{EAGLE-3 chain} & \textbf{EAGLE-3 tree} \\
\midrule
General VQA & \ensuremath{-}2.02 \ensuremath{\pm}0.32 & +1.37 \ensuremath{\pm}0.26 & +1.79 \ensuremath{\pm}0.06 & \ensuremath{-}4.44 \ensuremath{\pm}0.44 & \ensuremath{-}15.07 \ensuremath{\pm}0.55 & \ensuremath{-}54.16 \ensuremath{\pm}0.17 \\
Text VQA & \ensuremath{-}3.79 \ensuremath{\pm}0.17 & +3.66 \ensuremath{\pm}0.63 & +3.96 \ensuremath{\pm}0.21 & \ensuremath{-}4.26 \ensuremath{\pm}0.27 & \ensuremath{-}11.63 \ensuremath{\pm}0.03 & \ensuremath{-}45.68 \ensuremath{\pm}0.24 \\
Image captioning & \ensuremath{-}0.31 \ensuremath{\pm}0.25 & +1.02 \ensuremath{\pm}0.48 & +2.04 \ensuremath{\pm}0.31 & \ensuremath{-}6.79 \ensuremath{\pm}0.97 & \ensuremath{-}17.35 \ensuremath{\pm}0.23 & \ensuremath{-}59.39 \ensuremath{\pm}0.02 \\
Chart understanding & +3.56 \ensuremath{\pm}0.21 & +5.86 \ensuremath{\pm}0.41 & +6.15 \ensuremath{\pm}0.64 & \ensuremath{-}4.26 \ensuremath{\pm}0.09 & \ensuremath{-}10.94 \ensuremath{\pm}0.24 & \ensuremath{-}54.37 \ensuremath{\pm}0.08 \\
Complex reasoning & +3.46 \ensuremath{\pm}0.63 & +4.36 \ensuremath{\pm}0.00 & +4.57 \ensuremath{\pm}0.11 & \ensuremath{-}5.60 \ensuremath{\pm}0.30 & \ensuremath{-}17.60 \ensuremath{\pm}0.72 & \ensuremath{-}57.96 \ensuremath{\pm}0.15 \\
Multi-turn conversation & +5.24 \ensuremath{\pm}0.01 & +10.38 \ensuremath{\pm}0.30 & +11.34 \ensuremath{\pm}0.25 & \ensuremath{-}4.82 \ensuremath{\pm}0.03 & \ensuremath{-}14.21 \ensuremath{\pm}0.27 & \ensuremath{-}47.44 \ensuremath{\pm}0.29 \\
\bottomrule
\end{tabularx}
\end{table}

\begin{table}[htbp]
\caption{Per-task E2E throughput change relative to DSpark at batch size 32.}
\label{tab:sourceC2}
\centering
\small
\setlength{\tabcolsep}{4pt}
\begin{tabularx}{\textwidth}{@{}>{\raggedright\arraybackslash}X r r r r r r@{}}
\toprule
\textbf{Task} & \textbf{VisHead-0} & \textbf{VisHead-8} & \textbf{VisHead-4} & \textbf{DFlash} & \textbf{EAGLE-3 chain} & \textbf{EAGLE-3 tree} \\
\midrule
General VQA & \ensuremath{-}1.13 \ensuremath{\pm}0.62 & \ensuremath{-}0.31 \ensuremath{\pm}0.93 & +0.21 \ensuremath{\pm}0.08 & \ensuremath{-}4.73 \ensuremath{\pm}0.01 & \ensuremath{-}21.29 \ensuremath{\pm}2.25 & \ensuremath{-}50.67 \ensuremath{\pm}0.11 \\
Text VQA & \ensuremath{-}5.05 \ensuremath{\pm}0.34 & +1.88 \ensuremath{\pm}0.53 & +2.09 \ensuremath{\pm}0.47 & \ensuremath{-}4.78 \ensuremath{\pm}0.17 & \ensuremath{-}16.56 \ensuremath{\pm}0.57 & \ensuremath{-}43.89 \ensuremath{\pm}0.04 \\
Image captioning & \ensuremath{-}0.04 \ensuremath{\pm}0.71 & +1.60 \ensuremath{\pm}0.04 & +0.65 \ensuremath{\pm}0.03 & \ensuremath{-}7.22 \ensuremath{\pm}0.37 & \ensuremath{-}20.19 \ensuremath{\pm}0.44 & \ensuremath{-}52.56 \ensuremath{\pm}0.45 \\
Chart understanding & +1.55 \ensuremath{\pm}1.03 & +3.64 \ensuremath{\pm}0.71 & +3.42 \ensuremath{\pm}0.44 & \ensuremath{-}4.16 \ensuremath{\pm}0.48 & \ensuremath{-}15.89 \ensuremath{\pm}0.09 & \ensuremath{-}49.87 \ensuremath{\pm}0.31 \\
Complex reasoning & +2.18 \ensuremath{\pm}0.41 & +2.49 \ensuremath{\pm}0.87 & +1.28 \ensuremath{\pm}0.10 & \ensuremath{-}6.65 \ensuremath{\pm}0.02 & \ensuremath{-}21.25 \ensuremath{\pm}0.12 & \ensuremath{-}52.59 \ensuremath{\pm}0.01 \\
Multi-turn conversation & +4.00 \ensuremath{\pm}0.69 & +7.53 \ensuremath{\pm}0.41 & +6.80 \ensuremath{\pm}0.32 & \ensuremath{-}5.53 \ensuremath{\pm}0.18 & \ensuremath{-}20.88 \ensuremath{\pm}0.40 & \ensuremath{-}44.62 \ensuremath{\pm}0.86 \\
\bottomrule
\end{tabularx}
\end{table}

\begin{table}[htbp]
\caption{Per-task E2E throughput change relative to DSpark at batch size 16.}
\label{tab:sourceC3}
\centering
\small
\setlength{\tabcolsep}{4pt}
\begin{tabularx}{\textwidth}{@{}>{\raggedright\arraybackslash}X r r r r r r@{}}
\toprule
\textbf{Task} & \textbf{VisHead-0} & \textbf{VisHead-8} & \textbf{VisHead-4} & \textbf{DFlash} & \textbf{EAGLE-3 chain} & \textbf{EAGLE-3 tree} \\
\midrule
General VQA & \ensuremath{-}1.46 \ensuremath{\pm}0.09 & +0.47 \ensuremath{\pm}0.54 & +0.11 \ensuremath{\pm}0.11 & \ensuremath{-}4.05 \ensuremath{\pm}0.56 & \ensuremath{-}22.15 \ensuremath{\pm}0.58 & \ensuremath{-}42.64 \ensuremath{\pm}0.22 \\
Text VQA & \ensuremath{-}6.30 \ensuremath{\pm}0.09 & +0.73 \ensuremath{\pm}0.95 & +1.82 \ensuremath{\pm}0.68 & \ensuremath{-}3.92 \ensuremath{\pm}0.77 & \ensuremath{-}16.24 \ensuremath{\pm}0.56 & \ensuremath{-}37.48 \ensuremath{\pm}0.45 \\
Image captioning & \ensuremath{-}0.71 \ensuremath{\pm}1.07 & +1.40 \ensuremath{\pm}0.24 & +0.20 \ensuremath{\pm}1.36 & \ensuremath{-}6.55 \ensuremath{\pm}0.99 & \ensuremath{-}17.80 \ensuremath{\pm}1.11 & \ensuremath{-}43.41 \ensuremath{\pm}0.39 \\
Chart understanding & \ensuremath{-}1.04 \ensuremath{\pm}0.47 & +2.00 \ensuremath{\pm}1.13 & +2.17 \ensuremath{\pm}0.78 & \ensuremath{-}5.00 \ensuremath{\pm}0.46 & \ensuremath{-}16.03 \ensuremath{\pm}0.02 & \ensuremath{-}42.56 \ensuremath{\pm}0.61 \\
Complex reasoning & \ensuremath{-}1.69 \ensuremath{\pm}0.60 & +0.19 \ensuremath{\pm}0.89 & \ensuremath{-}0.22 \ensuremath{\pm}0.13 & \ensuremath{-}7.21 \ensuremath{\pm}1.22 & \ensuremath{-}16.06 \ensuremath{\pm}0.28 & \ensuremath{-}42.52 \ensuremath{\pm}0.45 \\
Multi-turn conversation & +0.73 \ensuremath{\pm}0.88 & +5.55 \ensuremath{\pm}0.58 & +5.74 \ensuremath{\pm}0.83 & \ensuremath{-}6.18 \ensuremath{\pm}0.41 & \ensuremath{-}14.12 \ensuremath{\pm}1.05 & \ensuremath{-}35.72 \ensuremath{\pm}0.84 \\
\bottomrule
\end{tabularx}
\end{table}

\begin{table}[htbp]
\caption{Per-task E2E throughput change relative to DSpark at batch size 8.}
\label{tab:sourceC4}
\centering
\small
\setlength{\tabcolsep}{4pt}
\begin{tabularx}{\textwidth}{@{}>{\raggedright\arraybackslash}X r r r r r r@{}}
\toprule
\textbf{Task} & \textbf{VisHead-0} & \textbf{VisHead-8} & \textbf{VisHead-4} & \textbf{DFlash} & \textbf{EAGLE-3 chain} & \textbf{EAGLE-3 tree} \\
\midrule
General VQA & \ensuremath{-}1.96 \ensuremath{\pm}0.37 & \ensuremath{-}1.18 \ensuremath{\pm}2.18 & \ensuremath{-}1.19 \ensuremath{\pm}1.01 & \ensuremath{-}4.20 \ensuremath{\pm}0.38 & \ensuremath{-}20.22 \ensuremath{\pm}0.94 & \ensuremath{-}25.96 \ensuremath{\pm}0.25 \\
Text VQA & \ensuremath{-}7.82 \ensuremath{\pm}0.21 & +0.36 \ensuremath{\pm}0.04 & +0.50 \ensuremath{\pm}0.37 & \ensuremath{-}5.43 \ensuremath{\pm}0.16 & \ensuremath{-}15.19 \ensuremath{\pm}0.01 & \ensuremath{-}22.75 \ensuremath{\pm}0.30 \\
Image captioning & \ensuremath{-}3.10 \ensuremath{\pm}2.22 & \ensuremath{-}0.28 \ensuremath{\pm}0.84 & \ensuremath{-}0.87 \ensuremath{\pm}2.42 & \ensuremath{-}6.94 \ensuremath{\pm}0.33 & \ensuremath{-}17.39 \ensuremath{\pm}0.31 & \ensuremath{-}22.88 \ensuremath{\pm}0.77 \\
Chart understanding & \ensuremath{-}2.41 \ensuremath{\pm}0.99 & +0.06 \ensuremath{\pm}0.53 & +0.56 \ensuremath{\pm}0.69 & \ensuremath{-}5.42 \ensuremath{\pm}1.03 & \ensuremath{-}13.90 \ensuremath{\pm}0.60 & \ensuremath{-}25.86 \ensuremath{\pm}0.07 \\
Complex reasoning & \ensuremath{-}0.32 \ensuremath{\pm}1.01 & \ensuremath{-}0.63 \ensuremath{\pm}0.09 & \ensuremath{-}0.45 \ensuremath{\pm}1.16 & \ensuremath{-}3.67 \ensuremath{\pm}0.44 & \ensuremath{-}13.67 \ensuremath{\pm}0.00 & \ensuremath{-}24.58 \ensuremath{\pm}0.08 \\
Multi-turn conversation & \ensuremath{-}0.31 \ensuremath{\pm}0.33 & +5.45 \ensuremath{\pm}0.39 & +3.46 \ensuremath{\pm}0.49 & \ensuremath{-}5.62 \ensuremath{\pm}0.35 & \ensuremath{-}11.38 \ensuremath{\pm}0.15 & \ensuremath{-}19.49 \ensuremath{\pm}0.12 \\
\bottomrule
\end{tabularx}
\end{table}

\begin{table}[htbp]
\caption{Per-task E2E throughput change relative to DSpark at batch size 1, using service times from the mixed runs.}
\label{tab:sourceC5}
\centering
\small
\setlength{\tabcolsep}{4pt}
\begin{tabularx}{\textwidth}{@{}>{\raggedright\arraybackslash}X r r r r r r@{}}
\toprule
\textbf{Task} & \textbf{VisHead-0} & \textbf{VisHead-8} & \textbf{VisHead-4} & \textbf{DFlash} & \textbf{EAGLE-3 chain} & \textbf{EAGLE-3 tree} \\
\midrule
General VQA & \ensuremath{-}1.68 \ensuremath{\pm}0.03 & \ensuremath{-}1.72 \ensuremath{\pm}0.47 & \ensuremath{-}0.67 \ensuremath{\pm}0.13 & \ensuremath{-}5.06 \ensuremath{\pm}0.54 & \ensuremath{-}18.32 \ensuremath{\pm}0.29 & +5.10 \ensuremath{\pm}1.41 \\
Text VQA & \ensuremath{-}10.82 \ensuremath{\pm}0.11 & \ensuremath{-}1.80 \ensuremath{\pm}0.20 & \ensuremath{-}2.55 \ensuremath{\pm}0.50 & \ensuremath{-}5.70 \ensuremath{\pm}0.78 & \ensuremath{-}19.63 \ensuremath{\pm}0.35 & +2.18 \ensuremath{\pm}0.46 \\
Image captioning & \ensuremath{-}1.60 \ensuremath{\pm}0.57 & \ensuremath{-}1.86 \ensuremath{\pm}0.19 & \ensuremath{-}1.23 \ensuremath{\pm}0.47 & \ensuremath{-}5.52 \ensuremath{\pm}0.05 & \ensuremath{-}18.74 \ensuremath{\pm}0.24 & +9.12 \ensuremath{\pm}0.03 \\
Chart understanding & \ensuremath{-}2.98 \ensuremath{\pm}0.37 & \ensuremath{-}1.62 \ensuremath{\pm}0.06 & \ensuremath{-}1.28 \ensuremath{\pm}0.27 & \ensuremath{-}4.87 \ensuremath{\pm}0.50 & \ensuremath{-}17.63 \ensuremath{\pm}0.13 & \ensuremath{-}0.80 \ensuremath{\pm}0.59 \\
Complex reasoning & \ensuremath{-}1.32 \ensuremath{\pm}0.35 & \ensuremath{-}1.39 \ensuremath{\pm}0.08 & \ensuremath{-}1.13 \ensuremath{\pm}0.45 & \ensuremath{-}5.30 \ensuremath{\pm}0.02 & \ensuremath{-}18.38 \ensuremath{\pm}0.01 & +2.03 \ensuremath{\pm}0.17 \\
Multi-turn conversation & \ensuremath{-}3.72 \ensuremath{\pm}0.44 & \ensuremath{-}0.07 \ensuremath{\pm}0.14 & \ensuremath{-}0.12 \ensuremath{\pm}0.22 & \ensuremath{-}5.84 \ensuremath{\pm}0.04 & \ensuremath{-}17.41 \ensuremath{\pm}0.46 & +4.16 \ensuremath{\pm}0.32 \\
\bottomrule
\end{tabularx}
\end{table}

\FloatBarrier
\vspace{1em}
\section{Additional Figures}
\label{app:sourceD}

\begin{figure}[htbp]
\centering
\includegraphics[width=0.76\linewidth]{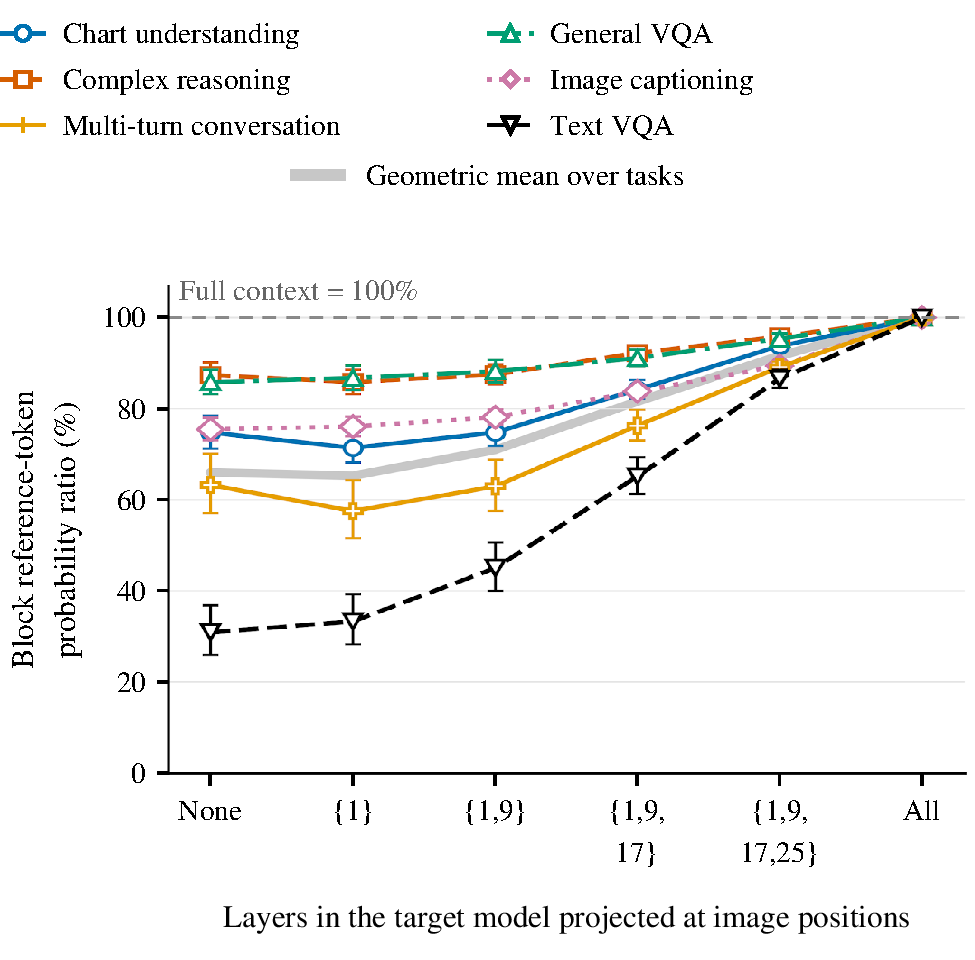}
\caption{Reference-token probability ratios when image rows are constructed from subsets of the sampled layers in the target model. We use Eq.~\eqref{eq:source1} with the subset-conditioned probability in place of \emph{q}\textsubscript{0}, averaged geometrically over the five draft offsets; full context is 100\%. Layers are added in increasing index. Tick labels list the selected layers; ``None'' removes image rows, and ``All'' uses \emph{S} = \{1, 9, 17, 25, 33\}. Text rows use all five layers throughout. Colored lines show the six tasks and error bars paired 95\% confidence intervals across prompts. Figure~\ref{fig:source4} shows the decreasing ordering.}
\label{fig:sourceD1}
\end{figure}

\begin{figure}[htbp]
\centering
\includegraphics[width=0.90\linewidth]{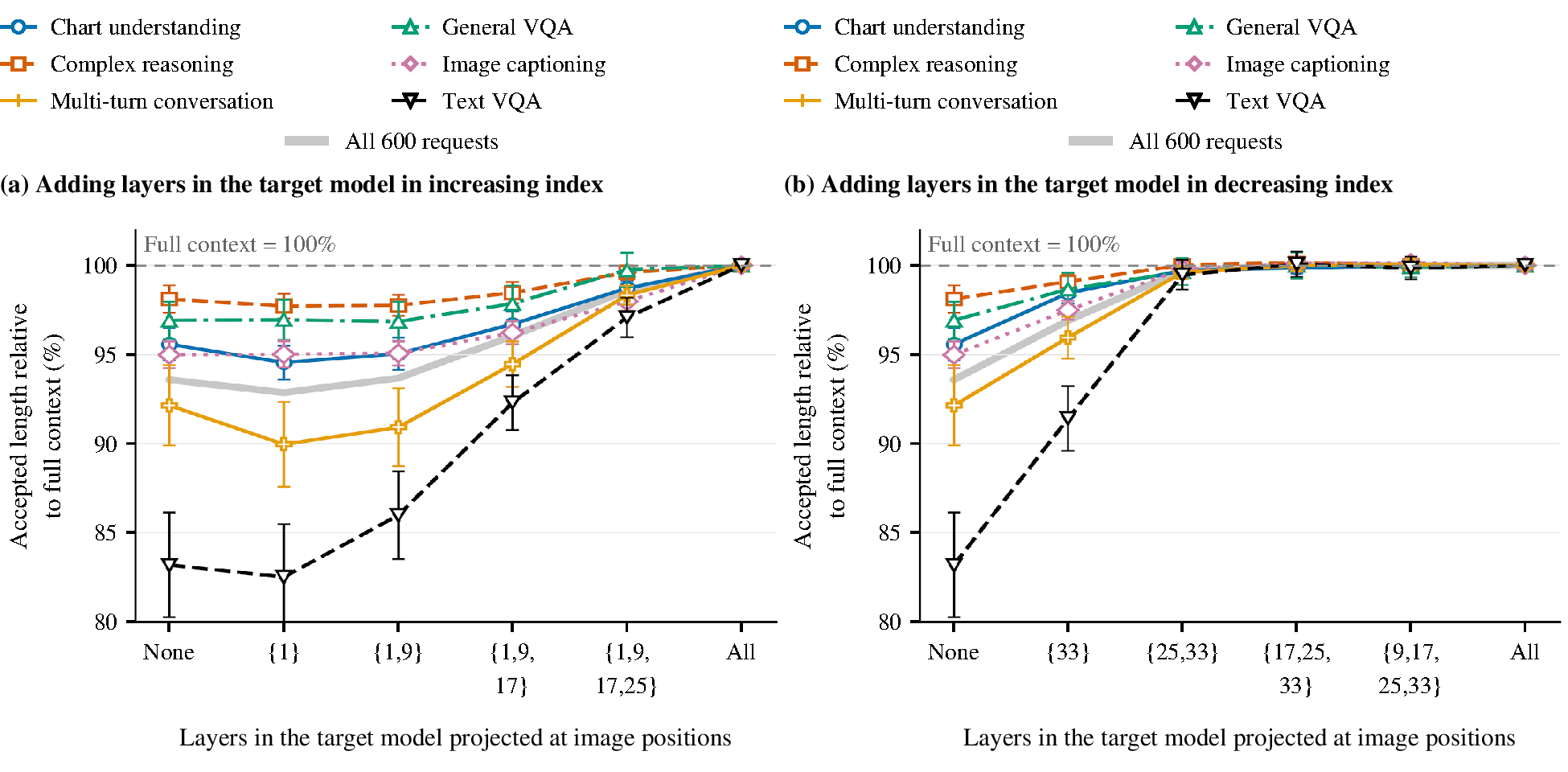}
\caption{Served accepted length when image rows use subsets of the sampled layers in the target model, relative to full context. Layers are added (a) in increasing and (b) in decreasing index; tick labels list selected layers. ``None'' removes image rows; ``All'' uses \emph{S} = \{1, 9, 17, 25, 33\}. Text rows use all five layers throughout.}
\label{fig:sourceD2}
\end{figure}

Figure~\ref{fig:sourceD2} shows that this probability retention carries over to accepted length in serving: using only layers 25 and 33 also gives served AL close to that obtained with all five sampled layers in the target model.

\begin{figure}[htbp]
\centering
\includegraphics[width=0.76\linewidth]{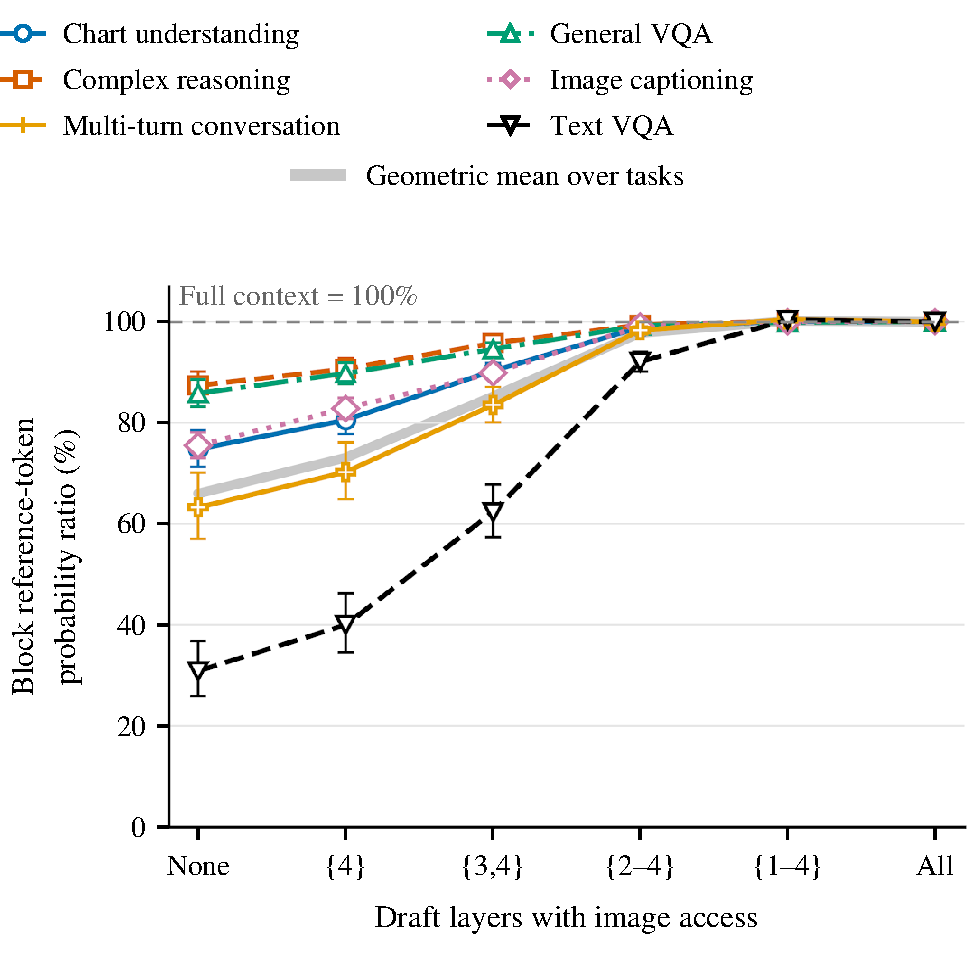}
\caption{Reference-token probability ratios when selected draft layers' KV heads may attend to image rows. Layers gain image access in decreasing index; tick labels list enabled layers, ending at full context. The gray line is the geometric mean over six tasks; error bars show paired 95\% confidence intervals across prompts. The increasing ordering is Figure~\ref{fig:source5}.}
\label{fig:sourceD3}
\end{figure}

\begin{figure}[htbp]
\centering
\includegraphics[width=0.90\linewidth]{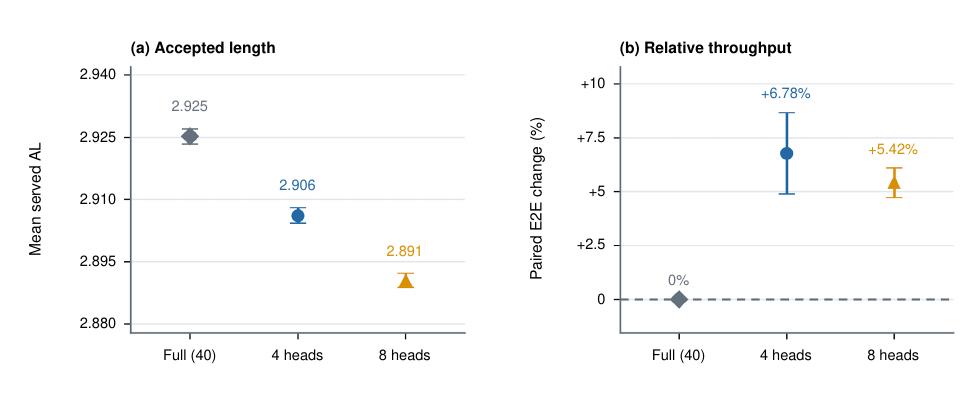}
\caption{Served AL (a) and paired E2E throughput change (b) for DFlash with full, four-head, and eight-head image access at a server batch cap of 64. Full-image DFlash is the throughput reference (0\%). Error bars are 95\% Student-t confidence intervals over three paired execution orders.}
\label{fig:sourceD4}
\end{figure}

\FloatBarrier
\vspace{1em}
\section{Compare dynamic routing with static routing}
\label{app:sourceE}

We compared a dynamic routing strategy for the KV head in the draft model with the static routing strategy mentioned in Section~\ref{sec:source3}. 

\textbf{Router design}. At inference the router may use only draft-side information. It makes an independent decision for each proposed draft tokens. Its inputs are the token's hidden state after the first two draft layers, which attend only to text, and a cached mean-pooled image embedding, which a rank-256 side network combines before a shared linear gate. The side network cannot modify the drafter's hidden state. Closed tokens attend to text keys and values only; open tokens read the image through the fixed VisHead-4 reader heads. We train the router on detector labels, calibrate it to 78\% visual recall on held-out data, and then fine-tune the router and reader heads jointly for one epoch.

\begin{figure}[htbp]
\centering
\includegraphics[width=0.90\linewidth]{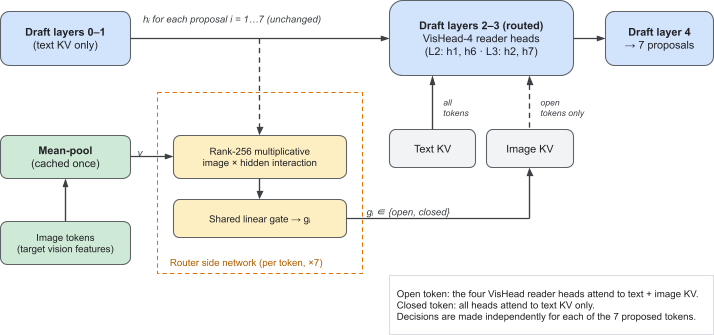}
\caption{the design of the head router}
\label{fig:sourceE1}
\end{figure}

\begin{table}[htbp]
\caption{Learned routing versus always-on VisHead-4. Top-1 agreement is on a held-out split; AL and throughput speedup are on MMSpec testmini with batch size 1.}
\label{tab:sourceE1}
\centering
\small
\setlength{\tabcolsep}{4pt}
\begin{tabularx}{\textwidth}{@{}>{\raggedright\arraybackslash}p{0.30\textwidth} >{\raggedright\arraybackslash}X r >{\centering\arraybackslash}p{0.16\textwidth}@{}}
\toprule
\textbf{Method} & \textbf{Top-1 agreement for  overall / visual / non-visual} & \textbf{AL} & \textbf{Throughput speedup} \\
\midrule
Full image access & 65.40 / 38.81 / 72.78 & 3.298 & 2.389\ensuremath{\times} \\
VisHead-4 (always on) & 64.33 / 34.87 / 72.50 & 3.240 & 2.271\ensuremath{\times} \\
VisHead-4 + learned per-token router & 64.10 / 34.05 / 72.43 & 3.187 & 2.219\ensuremath{\times} \\
\bottomrule
\end{tabularx}
\end{table}

The learned router does not improve on always-on VisHead-4 in this batch-1 comparison (Table~\ref{tab:sourceE1}). The reported changes are a decrease of 0.8 points in visual top-1 agreement, a 2.4\% decrease in AL, and a 2.3\% decrease in throughput. In this setting, the additional routing decisions sacrifice acceptance without improving throughput; the static mask avoids those per-token decisions.

\end{document}